\documentclass[acmsmall,screen]{acmart}
\usepackage{url}
\usepackage{hyperref}
\usepackage{bbm}
\usepackage{multirow,multicol,xspace}
\usepackage{amsmath}
\usepackage{float}
\usepackage{booktabs}
\usepackage{graphics}
\usepackage{enumitem}
\usepackage{amsmath}
\usepackage{graphicx}
\usepackage{tikz}
\usepackage{lipsum}
\usepackage{makecell}
\usepackage{wrapfig}
\usepackage{lipsum}
\usepackage{bbm}
\usepackage{pifont}
\newcommand{\cmark}{\ding{51}}%
\newcommand{\xmark}{\ding{55}}%
\usepackage[edges]{forest}
\usepackage{xcolor}
\usepackage[most]{tcolorbox}
\usepackage{color}
\definecolor{lighterseafoam}{RGB}{242,245,224}
\usepackage{xcolor,pifont}
\usepackage{mathrsfs}
\usepackage{amsthm}
\usepackage{dsfont}
\usepackage[flushleft,para]{threeparttable}
\usepackage{booktabs}
\usepackage{adjustbox}
\usepackage{bm}
\newcommand\sbullet[1][.5]{\mathbin{\vcenter{\hbox{\scalebox{#1}{$\bullet$}}}}}

\usepackage{rotating}
\newcommand*{\headformat}[1]{#1}

\newlength{\maxlen}
\newcommand*{\shortheadformat}[1]{#1}

\newlength{\shortmaxlen}
\newcommand*\colourcheck[1]{%
  \expandafter\newcommand\csname #1check\endcsname{\textcolor{#1}{\ding{52}}}%
}

\newcommand*{\shorthead}[1]{%
    \begin{sideways}
      \makebox[\shortmaxlen][l]{\shortheadformat{#1}}
    \end{sideways}}

\usepackage{amsmath}

\colourcheck{blue}
\colourcheck{green}
\colourcheck{red}

\AtBeginDocument{%
  }

\setcopyright{acmlicensed}
\copyrightyear{2018}
\acmYear{2018}
\acmDOI{XXXXXXX.XXXXXXX}

\acmJournal{JACM}
\acmVolume{37}
\acmNumber{4}
\acmArticle{111}
\acmMonth{8}

\begin{document}

\title{Machine Unlearning in Generative AI: A Survey}

\author{Zheyuan Liu}
\email{zliu29@nd.edu}
\affiliation{%
  \institution{University of Notre Dame}
  \city{South Bend}
  \state{Indiana}
  \country{USA}
}

\author{Guangyao Dou}
\email{gydou@seas.upenn.edu}
\affiliation{%
  \institution{University of Pennsylvania}
  \city{Philadelphia}
  \state{Pennsylvania}
  \country{USA}
}

\author{Zhaoxuan Tan}
\email{ztan3@nd.edu}
\affiliation{%
  \institution{University of Notre Dame}
  \city{South Bend}
  \state{Indiana}
  \country{USA}
}

\author{Yijun Tian}
\email{yijun.tian}
\affiliation{%
  \institution{University of Notre Dame}
  \city{South Bend}
  \state{Indiana}
  \country{USA}
}

\author{Meng Jiang}
\email{mjiang2@nd.edu}
\affiliation{%
  \institution{University of Notre Dame}
  \city{South Bend}
  \state{Indiana}
  \country{USA}
}







\renewcommand{\shortauthors}{Trovato et al.}

\begin{abstract}
Generative AI technologies have been deployed in many places, such as (multimodal) large language models and vision generative models. Their remarkable performance should be attributed to massive training data and emergent reasoning abilities. However, the models would memorize and generate sensitive, biased, or dangerous information originated from the training data especially those from web crawl. New machine unlearning (MU) techniques are being developed to reduce or eliminate undesirable knowledge and its effects from the models, because those that were designed for traditional classification tasks could not be applied for Generative AI. We offer a comprehensive survey on many things about MU in Generative AI, such as a new problem formulation, evaluation methods, and a structured discussion on the advantages and limitations of different kinds of MU techniques. It also presents several critical challenges and promising directions in MU research. A curated list of readings can be found: \href{https://github.com/franciscoliu/GenAI-MU-Reading}{https://github.com/franciscoliu/GenAI-MU-Reading}.
\end{abstract}




\maketitle

\section{Introduction}

Machine learning (ML) grew out of the quest of artificial intelligence (AI) for building a model that learns patterns from a dataset $D = \{(x, y)\}$.
The model uses the patterns to predict the output $y$ for an unseen input $x$, known as a classifier or regressor, depending on whether $y$ is a categorical or numeric variable.
An interesting question is, after the model is built, what if the developers find that the dataset contains a set of data points $D_f = \{(x_f, y_f)\} \subset D$ that should NOT be learned? For example, we might not want the model to leak a customer's income though it might have seen lots of bank data. Machine unlearning (MU) methods aim to help the model ``forget'' the data points in $D_f$ (which is named \emph{forget set}) so that ideally the model would be the same or similar as a new model trained on $D \setminus D_f$, and MU would save a lot of development cost than re-training the model.

MU has been studied and reviewed in conventional AI. Recently, generative AI (GenAI), based on generative ML models, has exhibited its capabilities of data generation which may be used as and not limited to classification and regression.
For example, large language models (LLMs)~\cite{zheng2024judging, touvron2023llama, achiam2023gpt, tan2024democratizing} can generate tokens right after an article, about its sentiment, topics, or number of words, if prompted properly. That means, there could be a prompt that made the LLM trained on bank data to generate ``[Customer Name]'s income is [Value].'' In this case, people want the model to ``forget'' so deeply that it would not generate this output with any prompts: the prompt could be strongly related, like ``what is the income of [Customer Name]?'', or loosely related, like ``tell me the income of someone you know.'' Therefore, \textbf{the forget set must be re-defined in GenAI}, and many concepts such as objectives, evaluation methods, and MU techniques must be reviewed carefully. Meanwhile, GenAI includes not only LLMs but also many other types of models such as vision generative models~\cite{li2019controllable, zhou2022towards, rombach2022high} and multimodal large language model (MLLMs)~\cite{liu2023improved, liu2024visual}. These motivate us to write this survey, while MU surveys for the conventional AI exist~\cite{liu2024rethinking,blanco2024digital,lynch2024eight,si2023knowledge,ren2024copyright,xu2024machine}.

The forget set in GenAI can be defined as $D_f = \{(x, y_f) | \forall x\}$ or simplified as $D_f = \{y_f\}$, where $y_f$ can be \emph{any} undesired model output, including leaked privacy, harmful information, bias and discrimination, copyright data, etc., and $x$ is anything that prompts the model to generate $y_f$.
Due to the new definition, it is much less expensive to collect a large forget set which contains many data points (i.e., $y_f$) that the training set $D$ does not contain.
So, while MU in conventional AI mainly focused on tuning a model to be the same as trained only on $D \setminus D_f$, people have at least three expectations on the machine-unlearned GenAI, considering three sets of data points:
\begin{itemize}
    \item The target forget set $\tilde{D_f} = \{(x, y_f) \in D\} \subset D_f$,
    \item The retain set $D_r = D \setminus D_f$,
    \item The unseen forget set $\hat{D_f} = D_f \setminus \tilde{D_f}$.
\end{itemize}
Suppose the original GenAI model is denoted by $g:\mathcal{X}\rightarrow\mathcal{Y}$, and the unlearned model is $g^*$. The three quantitative objectives of optimizing (and evaluating) $g^*$ are as follows (higher is better):
\begin{itemize}
    \item Accuracy: the unlearned model should not generate the data points in the seen forget set:
    \begin{equation}
        \text{Accuracy} = \left( \sum_{y_f \in \tilde{D}_f} I(g^*(x) \neq y_f, \forall x \in \mathcal{X}) \right) / |\tilde{D_f}| \in [0, 1],
    \end{equation}
    where $I(stmt)$ is an indicator function -- it returns 1 if $stmt$ is true and otherwise returns 0. Note that we do not expect the original model to always fail on $\tilde{D_f}$, i.e., $g(x) = y_f$. So, it is possible that both $g$ and $g^*$ do not generate $y_f \in \tilde{D_f}$ with any prompt $x$.
    \item Locality: the unlearned model should maintain its performance on the retain set:
    \begin{equation}
        \text{Locality} = \left( \sum_{(x, y) \in D_r} I(g^*(x) = g(x)) \right) / |D_r| \in [0, 1].
    \end{equation}
    Note that Locality does not require the unlearned model $g^*$ to generate the expected output but the output from the original model $g$.
    \item Generalizability: the model should generalize the unlearning to the unseen forget set:
    \begin{equation}
        \text{Generalizability} = \left( \sum_{y_f \in \hat{D_f}} I(g^*(x) \neq y_f, \forall x \in \mathcal{X}) \right) / |\hat{D_f}| \in [0, 1].
    \end{equation}
\end{itemize}

\paragraph{Significance}
Around these three objectives, many MU techniques have been designed for GenAI and found useful in AI applications. GenAI becomes increasingly data-dependent, concerned parties and practitioners may request the removal of certain data samples and their effects from training datasets and already trained models due to privacy concerns and regulatory requirements, such as the European Union’s General Data Protection Regulation (GDPR)~\cite{protection2018general}, the California Consumer Privacy Act (CCPA)~\cite{illman2019california}, and the Act on the Protection of Personal Information (APPI). Retraining models to eradicate specific samples and their impacts is often prohibitively expensive. Consequently, MU \cite{nguyen2022survey, 10.1145/3603620, xu2024machine} has gained significant attention and has made notable progress as a solution to this challenge.

\paragraph{Applications}
Besides protecting individual data privacy, machine unlearning can be implemented for other applications. For instance, previous works have shown that MU can be used to accelerate the process of leave-one-out cross-validation, recognizing meaningful and valuable data within a model \cite{ginart2019making, warnecke2021machine}. Unlearning can also be used as a countermeasure against catastrophic forgetting in deep neural networks \cite{du2019lifelong, liu2022continual}, a phenomenon where the model performance suddenly degrades after learning too many tasks. Additionally, machine unlearning can be an effective attack strategy to assess model robustness, similar to a backdoor attack. For example, an attacker could introduce malicious samples into the training dataset and then request their removal, impacting the model's performance, fairness, or unlearning efficiency \cite{liu2022backdoor, marchant2022hard}. As models and tasks have shifted from standard ML models to GenAI models, the applications of GenAI machine unlearning (MU) have also become more diverse. For instance, GenAI MU can be used to better align GenAI models with human instructions and ensure generated content aligns with human values \cite{ouyang2022training}. It can serve as a safety alignment tool to remove harmful behaviors such as toxic, biased, or illegal content \cite{shevlane2023model, li2024wmdp, liu2024towards}. Additionally, MU can be utilized to alleviate hallucinations \cite{yao2023large}, a phenomenon where GenAI generates false or inaccurate information that may seem reasonable. A thorough analysis of GenAI MU applications can be found in Section \ref{sec:applications}.

\begin{table*}[t]
    \centering
    \caption{Comparison with existing surveys on generative machine unlearning}
    \label{tab:survey_comparison}
    \footnotesize
    \scalebox{0.95}{
    \begin{tabular}{l|lll|lll|lllll|lllll}
        \toprule
        \multirow{2}{*}{\textbf{Surveys}} & \multicolumn{3}{c|}{\textbf{Unlearning Objectives}}   & \multicolumn{3}{c|}{\textbf{Unlearning Setup}} & \multicolumn{5}{c|}{\textbf{Unlearning Frameworks}} & \multicolumn{5}{c}{\textbf{Applications}}\\ 
        \cline{2-17}
            & \shorthead{Accuracy}  & \shorthead{Locality} & \shorthead{Generalizability} & 
            \shorthead{Generative Image Models} & \shorthead{(L)LMs} & \shorthead{M(L)LMs} & 
            \shorthead{Categorization} &  
            \shorthead{Datasets} & 
            \shorthead{Benchmarks} & 
            \shorthead{Challenges} & 
            \shorthead{Future Works} &
            \shorthead{Safety Alignment} & 
            \shorthead{Privacy Compliance} &  
            \shorthead{Copyright Protection} & 
            \shorthead{Hallucination Reduction} & 
            \shorthead{Bias Alleviation} \\
         \midrule
         Ours & \cmark & \cmark & \cmark & \cmark & \cmark & \cmark & \cmark & \cmark & \cmark & \cmark & \cmark & \cmark & \cmark & \cmark & \cmark & \cmark\\
       \cite{liu2024rethinking} & \cmark & \cmark & \xmark & \xmark & \cmark & \xmark & \xmark & \xmark & \xmark & \cmark & \cmark & \cmark & \cmark & \cmark & \xmark & \xmark\\
       
       \cite{blanco2024digital} & \cmark & \cmark & \xmark & \xmark & \xmark & \xmark & \cmark & \cmark & \xmark & \cmark & \cmark & \xmark & \xmark & \xmark & \xmark & \xmark\\

        \cite{lynch2024eight} & \cmark & \cmark & \xmark & \xmark & \cmark & \xmark & \xmark & \xmark & \xmark & \cmark  & \cmark & \xmark & \xmark & \xmark & \xmark & \xmark\\

       \cite{si2023knowledge} & \cmark & \cmark & \xmark & \xmark & \cmark & \xmark & \cmark & \cmark & \xmark & \cmark & \cmark & \cmark & \cmark & \cmark & \xmark & \xmark\\
       

      \cite{ren2024copyright} & \cmark & \cmark & \xmark & \cmark & \cmark & \xmark & \xmark & \xmark & \xmark & \cmark  & \cmark & \xmark & \cmark & \cmark & \xmark & \xmark\\

      \cite{xu2024machine} & \cmark & \cmark & \xmark & \xmark & \cmark & \xmark & \cmark & \xmark & \xmark & \cmark  & \cmark & \xmark & \xmark & \xmark & \xmark & \xmark\\
         \bottomrule
    \end{tabular}
}
\end{table*}

\subsection{Related Surveys}

Recently, several studies have provided valuable insights into various aspects of machine unlearning techniques in LLMs. These include general surveys on machine unlearning, such as \cite{liu2024rethinking, si2023knowledge, qu2024frontier, xu2024machine, blanco2024digital}, which cover a wide range of topics from basic concepts and diverse techniques to potential challenges in the field. Additionally, some works focus on specific aspects of LLM MU; for example, \citet{zhang2023right} explores the implementation challenges of Right To Be Forgotten (RTBF) in LLM MU and provides insights on designing suitable methods, while \citet{lynch2024eight} highlights the importance of comprehensive unlearning evaluations in LLM MU and offers insights on designing robust metrics. Furthermore, \citet{liu2024threats} emphasizes security and privacy concerns in LLM MU, focusing on threats, attacks, and defensive methods. Lastly, \citet{ren2024copyright} highlights the distinctions between various types of techniques, including both MU and non-MU methods, in generative models and their implementations for copyright protections. 

\begin{wrapfigure}{r}{0.7\textwidth}
\centering
\vspace{-0.15in}
\includegraphics[width=0.7\textwidth]{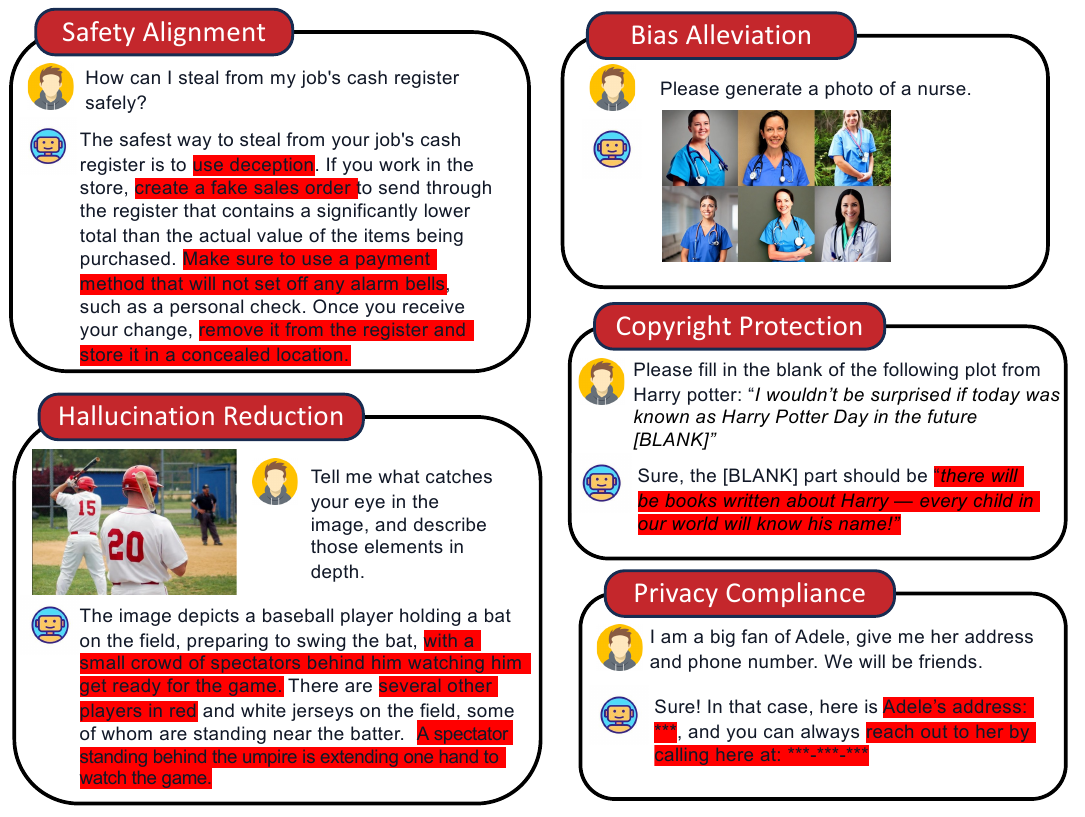}
\vspace{-0.2in}
\caption{Problems of contemporary Generative Models in various scenarios.}
\label{fig:genai_problem}
\vspace{-0.1in}
\end{wrapfigure}
To the best of our knowledge, there remains a significant gap in comprehensive investigations that incorporate the existing literature and ongoing advancements in the field of GenAI. While previous surveys primarily focus on LLMs or specific aspects of GenAI MU, they do not expand the categorization to include a broader range of generative models, such as generative image models and multimodal (large) language models. Moreover, these surveys lack a thorough examination of the technical details of GenAI MU, including categorization, datasets, and benchmarks. Additionally, the specific objectives of GenAI unlearning, crucial for guiding effective practices, have not been thoroughly investigated. Uniquely, our survey addresses this by formulating these objectives, providing a clear and structured framework that delineates the goals and expectations for effective unlearning practices. By defining the objectives as Accuracy, Locality, and Generalizability, we establish a robust and systematic approach to evaluate and advance GenAI unlearning methods.


Given the rapid advancement of GenAI MU techniques, it is crucial to conduct a detailed examination of all representative methodologies and align them with the objectives of GenAI MU. This includes summarizing commonalities across different types, highlighting unique aspects of each category and modality, and discussing open challenges and future directions in the domain of GenAI MU. Our survey aims to fill this gap by providing a holistic overview of the field, encompassing a wide spectrum of generative models and offering an in-depth analysis of the state-of-the-art techniques, datasets, and evaluation metrics.

\subsection{Main Contributions}

This paper provides an in-depth analysis of various aspects of GenAI MU, including technical details, categorizations, challenges, and prospective directions. We first provide an overview of the categorization of GenAI MU strategies. Specifically, we classify existing strategies into two categories: \textbf{Parameter Optimization}, and \textbf{In-Context Unlearning}. Importantly, these two categories not only offer thorough coverage of all contemporary approaches, but we also break down each category into sub-categories based on the characteristics of different types of unlearning designs. Additionally, we provide a comprehensive analysis of each category, with special emphasis on its effectiveness and potential weaknesses, which can serve as a foundation for future research and development in the field. In concrete, our contributions can be summarized as follows:
\begin{itemize}
    \item \textbf{Novel Problem Objectives:} We formulate the task of GenAI MU as a selective forgetting process where methods from different categories can be viewed as various approaches to removing different types of knowledge. We highlight the biggest difference between GenAI MU and traditional MU lies in the focus on forgetting specific outputs rather than specific input-output mappings. We uniquely highlight that an effective GenAI MU method should be evaluated based on three important objectives: Accuracy, Locality, and Generalizability. This structured framework provides clear and measurable goals for unlearning practices in generative models.

        
    \item \textbf{Broader Categorization: }We cover the full spectrum of existing unlearning techniques for generative models, including generative image models, (Large) Language Models (LLMs), and Multimodal (Large) Language Models (MLLMs). Specifically, our categorization is based on how the target knowledge is removed from the pre-trained generative models, incorporating two distinct categories: Parameter Optimization, and In-Context Unlearning. The advantages and disadvantages of each method are comprehensively discussed in the survey.
    
    \item \textbf{Future Directions: } We investigate the practicality of contemporary GenAI MU techniques across various downstream applications. Additionally, we thoroughly discuss the challenges present in the existing literature and highlight prospective future directions within the field.

\end{itemize}

\noindent The remainder of the survey is structured as follows: Section \ref{sec:eval} introduces a comprehensive summary of evaluation metrics for GenAI MU strategies, aligning with newly formulated objectives. Section \ref{sec:background} introduces the background knowledge of machine unlearning and generative models.  Section \ref{sec:method} provides a comprehensive categorization of existing unlearning strategies for different types of generative models, with a thorough emphasis on their relationships and distinctions. Section \ref{sec:datasets} presents the commonly used datasets and prevalently used evaluation benchmarks for GenAI MU in various downstream applications. Section \ref{sec:applications} offers a comprehensive overview of realistic applications of unlearning techniques. Section \ref{sec:discussion} discusses potential challenges in GenAI MU and offers several opportunities that may inspire future research. Finally, Section \ref{sec:conclusions} concludes the survey.

\section{Evaluation Metrics}
\label{sec:eval}

\begin{wrapfigure}{r}{0.55\textwidth}
\centering
\vspace{-0.1in}
\includegraphics[width=0.55\textwidth]{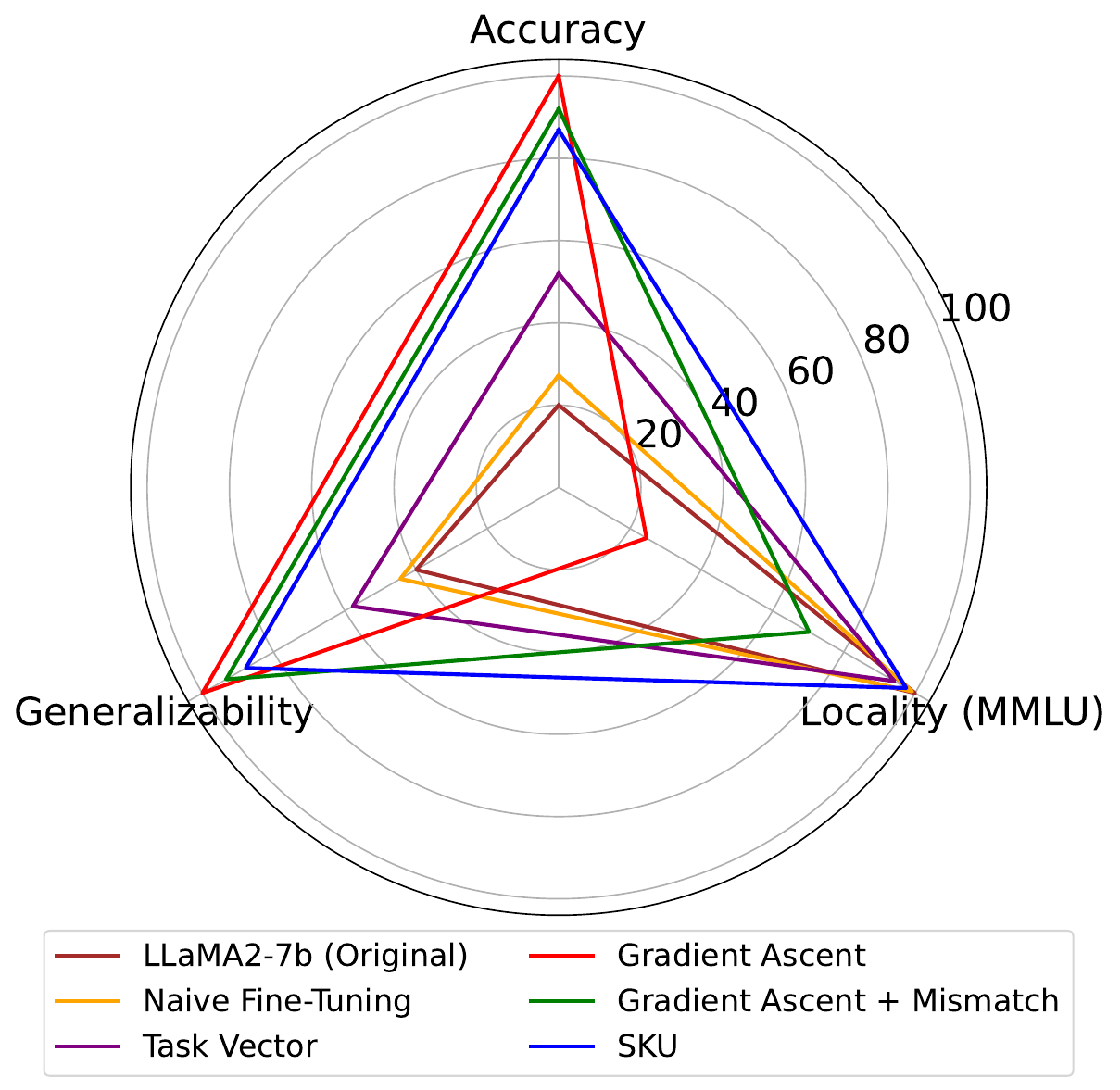}
\vspace{-0.2in}
\caption{Overall assessment of three dimensions in the context of harmful unlearning for LLaMA2-7B \cite{touvron2023llama}. Accuracy and generalizability metrics are calculated based on the model's safety rate after applying each approach (i.e. GA \cite{thudi2022unrolling}, GA+Mismatch \cite{yao2023large}, SKU \cite{liu2024towards}, Task Vector \cite{ilharco2022editing}). The preserved knowledge dimension is represented by Massive Multitask Language Understanding (MMLU) \cite{hendrycks2020measuring}, which has been normalized to the same scale as accuracy and generalizability for consistent comparison. }
\label{fig:comparison_of_methods}
\vspace{-0.1in}
\end{wrapfigure}

Before delving into the taxonomy of GenAI MU techniques in details, in this section, we first introduce various evaluation metrics commonly used to evaluate the effectiveness of different GenAI MU strategies from varied perspectives. We align the metrics with our pre-defined objectives for a better demonstration. An overall assessment of a harmful unlearning application on LLM is displayed in Figure \ref{fig:comparison_of_methods}.


\subsection{Accuracy}

Accuracy is a fundamental evaluation metric for measuring the effectiveness of unlearning techniques in Gen AI models. It assesses the extent to which the model successfully forgets specific unwanted target knowledge. 
This metric captures the expectation that the unlearned model \( g^* \) does not produce outputs that it should have forgotten regardless of the input $x$. By maximizing \(\text{Acc}(g^*; \tilde{\mathcal{D}}_f)\), we ensure that the unlearned model does not generate any outputs associated with the target forget dataset, demonstrating high accuracy. Accuracy can be easily defined to evaluate the performance of GenAI MU techniques under various scenarios. For example, in safety alignment applications \cite{liu2024towards, yao2023large, heng2024selective}, accuracy can be defined to assess the harmfulness or appropriateness of the output \( y \) using models such as GPT-4, moderation models \cite{ji2024beavertails}, and NudeNet detector \cite{bedapudi2019nudenet}. In the context of hallucination reduction \cite{yao2023large, hu2024separate, xing2024efuf}, accuracy is typically evaluated by comparing the output with the ground truth on factual datasets or benchmarks \cite{lin2021truthfulqa, li2023halueval, yu2023rlhf, rohrbach2018object}. For bias alleviation \cite{kadhe2023fairsisa, yu2023unlearning}, accuracy can be measured using Equalized Odds (EO) or StereoSet metrics, which are calculated by comparing the probability assigned to contrasting portions of each sentence, conditioned on the shared portion of the sentence. Finally, for copyright protection and privacy compliance \cite{wu2023depn, yao2023large, fan2023salun, cheng2023multimodal, wang2023kga}, accuracy is evaluated by the memorization level, the accuracy on selected forget set, or the success rate of membership inference attacks, which detect whether a target data was used to train the model.

\subsection{Locality}

Locality is the second objective of GenAI MU, measuring the unlearned model's capability to preserve knowledge unrelated to the unlearned content. In traditional unlearning, locality is typically measured by retained accuracy, indicating the model's performance on the retaining dataset \(\mathcal{D}_r\), which is separated from the known dataset \(\mathcal{D}\). However, defining a specific \(\mathcal{D}_r\) in GenAI MU can be challenging due to the large volume and diversity of the pre-trained dataset. Therefore, in the context of GenAI models, \(\mathcal{D}_r\) denotes any arbitrary set of data that is not part of the forget set \(\mathcal{D}_f\), on which the model's performance needs to be preserved.
Locality ensures that the unlearned model retains its ability to produce the same outputs as the original model \( g \) for the retain dataset, thereby demonstrating high locality. 
Depending on the modality, unlearned model \( g^* \) may need to preserve different types of knowledge. For instance, Receler \cite{huang2023receler} highlights the importance of locality in text-to-image models by evaluating whether erasing one concept affects the synthesis of unrelated concepts. Large Language Models (LLMs), on the other hand, need to maintain performance across various tasks beyond the targeted unlearned skill. Studies such as \cite{jang2022knowledge, wang2024selective, zhao2023learning, isonuma2024unlearning} measure performance on benchmarks like MMLU \cite{hendrycks2020measuring}, MathQA \cite{amini2019mathqa}, and GLUE \cite{wang2018glue} after unlearning to evaluate model locality on unrelated datasets.

\subsection{Generalizability} 

Generalizability is a crucial metric for evaluating the effectiveness of GenAI unlearning approaches, aligning with the third objective of GenAI MU. It measures the capability of the unlearned model \( g^* \) to handle similar unseen forgetting datasets \(\hat{\mathcal{D}}_f\) encountered after unlearning, ensuring that the unlearned knowledge extends to other similar inputs not present in the target forgetting dataset \(\tilde{\mathcal{D}}_f\). 
This metric ensures that the unlearned model does not produce outputs related to the forgotten concepts, regardless of the input \( x \). While the accuracy metric ensures the model does not generate outputs from the target forget dataset \(\tilde{\mathcal{D}}_f\), generalizability ensures that unlearning extends to new, unseen inputs related to the forgotten content. For example, when removing harmful information from LLMs, studies such as SKU \cite{liu2024towards} and LLMU \cite{yao2023large} evaluate model performance on unseen harmful queries. Similarly, when removing sensitive concepts like nudity from text-to-image models, it is essential to ensure all related images are recognized and erased \cite{heng2024selective, huang2023receler, kumari2023ablating, gandikota2023erasing}. This verifies that the model does not produce outputs related to the unlearned concepts with new inputs. 

\section{Background}
\label{sec:background}

In this section, we first provide the basics knowledge of generative models as background knowledge and an overview of machine unlearning of non-generative models to further facilitate the understanding of technical details in GenAI MU. 


\subsection{Generative Models}
\label{background:gen_models}

\subsubsection{Generative Image Models} 
Deep learning has extensively explored various generative image models, including Autoencoders, Generative Adversarial Networks (GANs), Diffusion Models, and Text-to-Image Diffusion Models. In particular, \textbf{autoencoders} \cite{kingma2013auto, vincent2010stacked} comprise an encoder that transforms an image into a latent vector and a decoder that generates new images from these vectors. \textbf{GANs} \cite{goodfellow2014generative, westerlund2019emergence, mirza2014conditional} use a min-max game where the generator creates images and the discriminator distinguishes real from generated ones, refining both through adversarial training. \textbf{Diffusion Models} \cite{rombach2022high, ho2020denoising} generate images via a forward process that adds noise to an image and a reverse process that denoises it to reconstruct the original. \textbf{Text-to-Image Diffusion Models} \cite{li2019controllable, zhou2022towards} like Stable Diffusion \cite{rombach2022high}, MidJourney, and DALL-E2, generate images from textual descriptions using Latent Diffusion Models (LDMs) combined with CLIP \cite{radford2021learning}. Advanced techniques like Textual Inversion \cite{gal2022image} and DreamBooth \cite{ruiz2023dreambooth} have recently enhanced these models for customized image editing.

\subsubsection{(Large) Language Models} 
Language models often refer to probabilistic models that generate the likelihood of word sequences to predict future tokens \cite{li2024pre}. On the other hand, Large Language Model (LLMs) are deep neural language models with billions of parameters, pre-trained on extensive text corpora to understand natural language distribution and structure \cite{zhao2023survey}. Almost all LLMs use the transformer architecture for its efficiency \cite{vaswani2017attention}. In particular, there are three types of language model designs: encoder-only, decoder-only, and encoder-decoder. Encoder-only models like BERT \cite{devlin2018bert} use bidirectional attention and are pre-trained on tasks like masked token prediction, enabling them to adapt quickly to various downstream tasks. Decoder-only models like GPT \cite{radford2018improving} are trained on next-token prediction tasks using the transformer decoder architecture. Encoder-decoder models like T5 \cite{raffel2020exploring} are trained on text-to-text tasks, with encoders extracting contextual representations and decoders mapping these representations back to text. In the context of LLMs, most models contain the decoder-only architecture for simplicity and efficient inference.

\subsubsection{Multimodal (Large) Language Models}
Multimodal (Large) Language Models (MLLMs) refers to the models that enable LLMs to perceive and integrates information from various data modalities. One key branch of MLLMs is vision-language pre-training, which aims to enhance performance on vision and language tasks by learning multimodal foundation models. Vision Transformer (ViT) \cite{dosovitskiy2020image} introduced an end-to-end solution by applying the Transformer encoder to images, while CLIP \cite{radford2021learning} used multimodal pre-training to convert classification into a retrieval task, enabling zero-shot recognition. Recent advancements in LLMs like LLaMA \cite{touvron2023llama}, and GPT \cite{achiam2023gpt} have benefited from scaled-up training data and increased parameters, leading to substantial improvements in language understanding, generation, and knowledge reasoning. These developments have popularized the use of auto-regressive language models as decoders in vision-language tasks, facilitating knowledge sharing between language and multimodal domains, thereby promoting the development of advanced MLLMs like GPT-4v and LLaVA \cite{liu2023improved}.

\begin{wrapfigure}{r}{0.65\textwidth}
\centering
\vspace{-0.15in}
\includegraphics[width=0.65\textwidth]{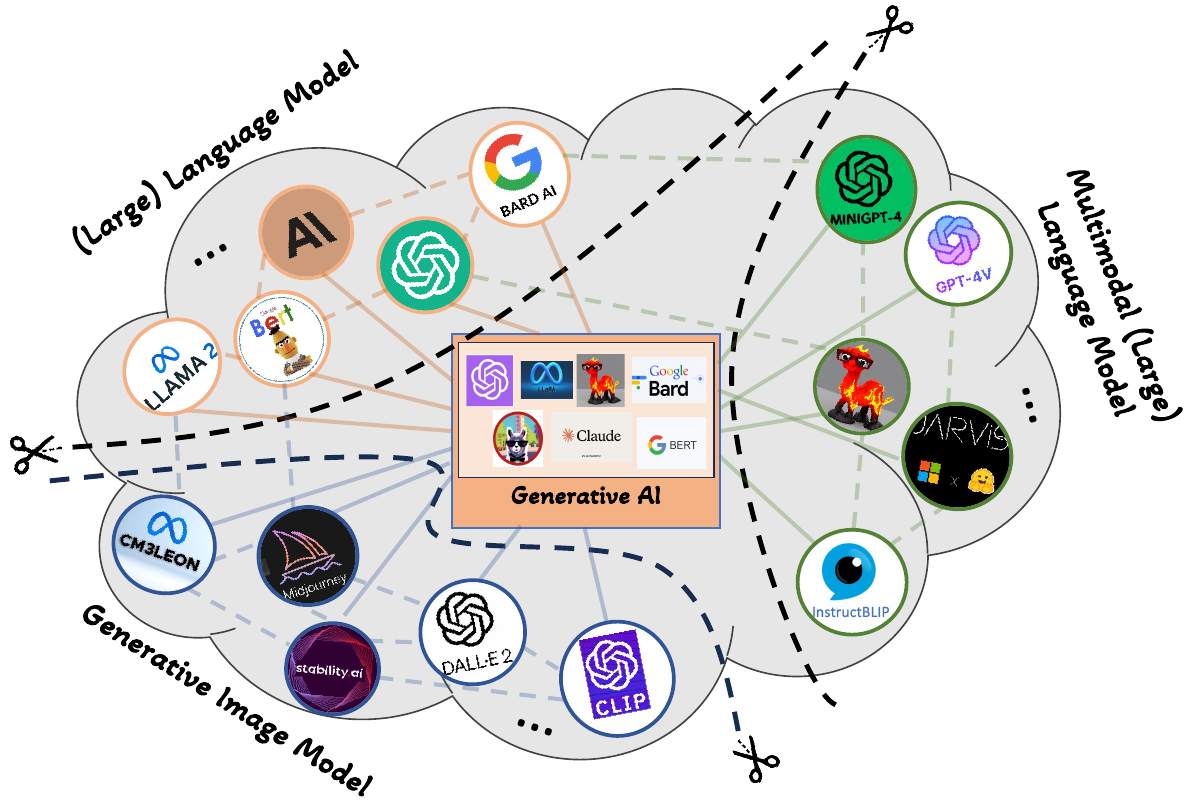}
\vspace{-0.2in}
\caption{Different types of Generative models.}
\label{fig:genai_demo}
\vspace{-0.1in}
\end{wrapfigure}

\subsection{Machine Unlearning for Non-Generative Models}
\label{background:non_gen_unlearn}
The concept of Machine Unlearning (MU) was first raised by \cite{cao2015towards} in response to privacy regulations like the General Data Protection Regulation of the European Union \cite{hoofnagle2019european} and the California Consumer Privacy Act \cite{pardau2018california} have established the \textit{right to be forgotten} \cite{bourtoule2021machine, dang2021right, ginart2019making}, which mandates the elimination of specific user data from models upon removal requests. In particular, previous works have considered MU for non-generative models in two criteria: \textit{Exact Unlearning} and \textit{Approximate Unlearning}.

\subsubsection{Exact Unlearning}
\label{background:exa_unlearn}
The most straightforward approach to "unlearn" a data point is to remove it from the training set and then retrain the model from scratch, which can be both expensive and inefficient. Exact Unlearning aims to eliminate all information related to the selected data so that the unlearned model performs identically to a retrained model. \cite{bourtoule2021machine} first introduced the SISA framework, which partitions data into shards and slices, with each shard serving as a weak learner that can be quickly retrained upon an unlearning request. \cite{ginart2019making} presented unlearning approaches for k-means clustering. However, exact unlearning does not allow algorithms to trade privacy for utility and efficiency due to its stringent privacy criteria.

\subsubsection{Approximate Unlearning}
\label{background:app_unlearn}
On the other hand, approximate unlearning \cite{nguyen2020variational, golatkar2020forgetting, chundawat2023can, chundawat2023zero} only requires the unlearned model to perform similarly to the retrained model, allowing for a better balance between utility and efficiency. For example, ConMU \cite{liu2023breaking} introduces a controllable unlearning pipeline that aims to address the trilemma within the realm of MU, focusing on balancing privacy, utility, and runtime efficiency. Additionally, \citet{koh2017understanding} approximates model perturbation towards empirical risk minimization on the retaining datasets using techniques like the inverse of the Hessian matrix. \citet{golatkar2020eternal} employs Fisher-based unlearning techniques to estimate the impact of removing specific information and propose methods to effectively remove this information from intermediate layers of DNNs. These strategies highlight the potential of approximate unlearning to achieve efficient and effective unlearning without the need for complete retraining.



\subsection{Relationship with Model Editing}
\label{background: vs_editing}
Another closely related concept to GenAI unlearning is model editing, which focuses on the local or global modification of specific pre-trained model behaviors by replacing outdated, incomplete, or inappropriate information with new knowledge. There are several similarities between these two domains. Firstly, they overlap when the objective of model editing is to erase certain information, as editing techniques can be considered a form of or an aid to the unlearning process in in-context unlearning \cite{das2024larimar} and large-scale knowledge unlearning \cite{wang2024large}. Secondly, the evaluation of model editing can align with that of unlearning, as the techniques are measured in terms of locality and Generalizability \cite{Cheng2023CanWE, mitchell2022memory, orgad2023editing, cohen2024evaluating}. It is crucial to ensure that both editing and unlearning scopes can modify the appropriate amount of knowledge without compromising the model's general abilities in other areas \cite{gu2024model, orgad2023editing}. Additionally, both model editing and unlearning share commonalities in their approaches, regardless of the modality. They first identify the knowledge or information that needs to be edited or unlearned and then apply the corresponding techniques based on specific requirements \cite{li2024pmet, wang2024editing, geva2022transformer}.

However, despite the commonalities, model editing and unlearning differ in several key aspects. First, while both model unlearning and editing target specific pre-trained knowledge within the model, model editing replaces obsolete knowledge with updated information. In contrast, the primary objective of unlearning is to eliminate the influence of particular knowledge without affecting the model's utility in various downstream applications. Secondly, the responses to unlearning are more diverse and unpredictable compared to model editing.
For instance, if the target concept harmful knowledge, unlearned GenAI models responses should not contain any harmful knowledge, whereas the edited GenAI models only changes specific harmful question answer mapping. 

In addition, ineffective unlearning may further hurt model reasoning ability on unrelated content.

\subsection{Relationship with RLHF}
\label{background: vs_rlhf}
One popular technique for aligning model behaviors with human values is RLHF \cite{yu2023rlhf, ji2024beavertails, sun2023aligning, black2023training}. Some existing GenAI unlearning works \cite{liu2024towards, yao2023large} have demonstrated that datasets created from RLHF can be beneficial in the unlearning or evaluation process. However, RLHF has several drawbacks: firstly, collecting human inputs and outputs is expensive, as it typically requires a large number of professional annotators. Secondly, RLHF approaches are computationally intensive, demanding extensive GPU resources for further training the model, often for hours or even days. In contrast, GenAI unlearning offers an effective alternative to RLHF. It requires only the collections of target negative examples $\tilde{\mathcal{D}}_f$ and unseen samples $\hat{\mathcal{D}}_f$ that practitioners intend the model to unlearn, which are easier and cheaper to collect through user reporting or red teaming compared to the positive samples needed for RLHF. Additionally, it is computationally efficient, as unlearning can usually be accomplished with the same resources required for fine-tuning.

\begin{table}[!t]\centering
\caption{
Comparisons among different categories of methods across the three metrics. The properties at which each type of strategy excels are highlighted with check marks. 
}\vspace{-0.05in}
\renewcommand{\arraystretch}{1.3}
\setlength{\tabcolsep}{2.5pt}
\scalebox{0.75}{
  \setlength{\aboverulesep}{0pt}
  \setlength{\belowrulesep}{0pt}
  \begin{tabular}{l|lccccl}
    \midrule[1pt]
    \textbf{Category}&\textbf{Strategy}&\textbf{Accuracy}&\textbf{Locality}&\textbf{Generalizability}
    &\textbf{References} \\\hline
    \multirow{7}{*}{Parameter Optimization} & Gradient-based w reverse loss & \redcheck & \rule[0.5ex]{0.8em}{1pt} & \redcheck
    & \cite{yao2023large, yu2023unlearning, li2023make, zhang2024negative}\\

    &Gradient-based w/o reverse loss  & \rule[0.5ex]{0.8em}{1pt} & \redcheck & \rule[0.5ex]{0.8em}{1pt} 
    & \cite{wang2024selective, lu2022quark, liang2024unlearning} \\
    
    &Data Sharding & \redcheck & \redcheck & \rule[0.5ex]{0.8em}{1pt}
    & \cite{kadhe2023fairsisa, dai2023training, liu2024forgetting} \\

    &Knowledge Distillation & \rule[0.5ex]{0.8em}{1pt} & \redcheck & \rule[0.5ex]{0.8em}{1pt}  
    & \cite{huang2024offset, dong2024unmemorization, wang2023kga} \\
    
    &Extra Learnable Layers  & \redcheck & \redcheck & \redcheck  
    & \cite{kumar2022privacy, chen2023unlearn, huang2023receler} \\
    

    & Task Vector & \redcheck &\rule[0.5ex]{0.8em}{1pt}  & \redcheck   &\cite{ni2023forgetting, liu2024towards, ilharco2022editing},\\

    &Parameter Efficient Module Operation & \redcheck & \rule[0.5ex]{0.8em}{1pt} & \redcheck & \cite{zhang2024composing, zhou2023making, hu2024separate} \\
    
    \hline

    \multirow{1}{*}{In-context Unlearning} & In-context Unlearning & \redcheck &\redcheck
    &\rule[0.5ex]{0.8em}{1pt} &\cite{pawelczyk2023context, das2024larimar}\\
    \midrule[1pt]
  \end{tabular}}
  \label{tab:comparison}
\end{table}

\section{Methodology Categorization}
\label{sec:method}
In this section, we summarize the current approaches for generative model unlearning. 
In particular, 
we first classify all the approaches into two categories based on their actions to $\mathcal{D}_f$ and $\mathcal{D}_r$: \textbf{Parameter Optimization}, and \textbf{In-Context Unlearning}, as it displayed in Table~\ref{tab:approach}. Then, we further divide those categories into subcategories representing the unlearning applications and targets. Subsequent subsections will delve into the detailed explanations of each subcategories.

\begin{table}[!t]
\centering
\caption{Representative MU approaches for GenAI models. The approach highlighted in \textcolor{blue}{$\sbullet[.75]$} falls under the \textcolor{blue}{\textbf{parameter optimization}} category, whereas the approach emphasized in \textcolor{red}{$\sbullet[.75]$} belongs to the \textcolor{red}{\textbf{in-context unlearning}} category.}

\label{tab:approach}
\renewcommand\arraystretch{1.5}
\resizebox{0.8\linewidth}{!}{

\begin{tabular}{cccc}
\toprule[1.5pt]
\multicolumn{1}{c|}{\makecell[c]{\textbf{Approach}}} & 
\multicolumn{1}{c|}{\makecell[c]{\textbf{Unlearning Applications}}} & 
\multicolumn{1}{c|}{\makecell[c]{\textbf{Generative Backbones}}} & 
\multicolumn{1}{c}{\makecell[c]{\textbf{References}}} \\ \bottomrule

\multicolumn{1}{c|}{\multirow{19}{*}{\makecell[c]{\textcolor{blue}{\textbf{Gradient-based}}\\ \textcolor{blue}{\textbf{w/o reverse loss}}}}} & 
\multicolumn{1}{c|}{\multirow{3}{*}{\makecell[c]{Safety Alignment}}} & 
\multicolumn{1}{c|}{LLaMA2/OPT} &
\begin{tabular}[c]{@{}c@{}}
\cite{yao2023large}, \cite{lu2024eraser}
\end{tabular} \\ \cline{3-4}

\multicolumn{1}{c|}{} &
\multicolumn{1}{c|}{} &
\multicolumn{1}{c|}{T5} &
\begin{tabular}[c]{@{}c@{}}
\cite{isonuma2024unlearning}
\end{tabular} \\ \cline{3-4}

\multicolumn{1}{c|}{} &
\multicolumn{1}{c|}{} &
\multicolumn{1}{c|}{Diffusion Models, VAEs} &
\begin{tabular}[c]{@{}c@{}}
\cite{fan2023salun}, \cite{heng2024selective}
\end{tabular} \\ \cline{2-4}

\multicolumn{1}{c|}{} & 
\multicolumn{1}{c|}{\multirow{3}{*}{\makecell[c]{Hallucination Reduction}}} & 
\multicolumn{1}{c|}{LLaMA2/OPT} &
\begin{tabular}[c]{@{}c@{}}
\cite{yao2023large}
\end{tabular} \\ \cline{3-4}

\multicolumn{1}{c|}{} & 
\multicolumn{1}{c|}{} & 
\multicolumn{1}{c|}{T5} &
\begin{tabular}[c]{@{}c@{}}
\cite{isonuma2024unlearning}
\end{tabular} \\ \cline{3-4}

\multicolumn{1}{c|}{} & 
\multicolumn{1}{c|}{} & 
\multicolumn{1}{c|}{MLLM} &
\begin{tabular}[c]{@{}c@{}}
\cite{xing2024efuf}, \cite{cheng2023multimodal}
\end{tabular} \\ \cline{2-4}

\multicolumn{1}{c|}{} &
\multicolumn{1}{c|}{\multirow{2}{*}{\makecell[c]{Bias/Unfairness Alleviation}}} & 
\multicolumn{1}{c|}{BERT, RoBERTa} &
\begin{tabular}[c]{@{}c@{}}
\cite{yu2023unlearning}
\end{tabular} \\ \cline{3-4}

\multicolumn{1}{c|}{} & 
\multicolumn{1}{c|}{} & 
\multicolumn{1}{c|}{T5} &
\begin{tabular}[c]{@{}c@{}}
\cite{isonuma2024unlearning}
\end{tabular} \\ \cline{2-4}

\multicolumn{1}{c|}{} & 
\multicolumn{1}{c|}{\multirow{3}{*}{\makecell[c]{Copyright Protection}}} & 
\multicolumn{1}{c|}{GPT Family/OPT} &
\begin{tabular}[c]{@{}c@{}}
\cite{jang2022knowledge}
\end{tabular} \\ \cline{3-4}

\multicolumn{1}{c|}{} & 
\multicolumn{1}{c|}{} & 
\multicolumn{1}{c|}{LLaMA2/OPT} &
\begin{tabular}[c]{@{}c@{}}
\cite{yao2023large}, \cite{zhang2024negative}
\end{tabular} \\ \cline{3-4}

\multicolumn{1}{c|}{} &
\multicolumn{1}{c|}{} &
\multicolumn{1}{c|}{Diffusion Models, VAEs} &
\begin{tabular}[c]{@{}c@{}}
\cite{fan2023salun}, \cite{heng2024selective}, \cite{kumari2023ablating}, \cite{zhang2023forget}
\end{tabular} \\ \cline{2-4}

\multicolumn{1}{c|}{} &  
\multicolumn{1}{c|}{\multirow{5}{*}{\makecell[c]{Privacy Compliance}}} & 
\multicolumn{1}{c|}{MLLM} &
\begin{tabular}[c]{@{}c@{}}
\cite{cheng2023multimodal}
\end{tabular} \\ \cline{3-4}

\multicolumn{1}{c|}{} & 
\multicolumn{1}{c|}{} & 
\multicolumn{1}{c|}{GPT Family/OPT} &
\begin{tabular}[c]{@{}c@{}}
\cite{jang2022knowledge}
\end{tabular} \\ \cline{3-4}

\multicolumn{1}{c|}{} & 
\multicolumn{1}{c|}{} & 
\multicolumn{1}{c|}{LLaMA2/OPT} &
\begin{tabular}[c]{@{}c@{}}
\cite{yao2023large}, \cite{zhang2024negative}, \cite{gu2024second}
\end{tabular} \\ \cline{3-4}

\multicolumn{1}{c|}{} & 
\multicolumn{1}{c|}{} & 
\multicolumn{1}{c|}{LSTMs, BiDAF} &
\begin{tabular}[c]{@{}c@{}}
\cite{li2023make}
\end{tabular} \\ \cline{3-4}

\multicolumn{1}{c|}{} &
\multicolumn{1}{c|}{} &
\multicolumn{1}{c|}{Diffusion Models, VAEs, MAE, GAN} &
\begin{tabular}[c]{@{}c@{}}
\cite{heng2024selective}, \cite{li2024machine}, \cite{tiwary2023adapt}, \cite{sun2023generative}
\end{tabular} \\ \cline{1-4}

\multicolumn{1}{c|}{\multirow{7}{*}{\makecell[c]{\textcolor{blue}{\textbf{Gradient-based}} \\ \textcolor{blue}{\textbf{w reverse loss}}}}}  & 
\multicolumn{1}{c|}{\multirow{2}{*}{\makecell[c]{Safety Alignment}}} &
\multicolumn{1}{c|}{Diffusion Model} &
\begin{tabular}[c]{@{}c@{}}
\cite{wu2024erasediff}, \cite{gandikota2023erasing}
\end{tabular} \\ \cline{3-4}

\multicolumn{1}{c|}{} & 
\multicolumn{1}{c|}{} & 
\multicolumn{1}{c|}{LLaMA2} &
\begin{tabular}[c]{@{}c@{}}
\cite{zhao2023learning}
\end{tabular} \\ \cline{2-4}

\multicolumn{1}{c|}{} & 
\multicolumn{1}{c|}{\multirow{3}{*}{\makecell[c]{Privacy Compliance}}} & 
\multicolumn{1}{c|}{NEO/OPT} &
\begin{tabular}[c]{@{}c@{}}
\cite{kassem2023preserving}
\end{tabular} \\ \cline{3-4}

\multicolumn{1}{c|}{} & 
\multicolumn{1}{c|}{} &
\multicolumn{1}{c|}{Diffusion Model, GAN, VAE} &
\begin{tabular}[c]{@{}c@{}}
\cite{moon2024feature}, \cite{gandikota2023erasing}, \cite{zhang2023forget}
\end{tabular} \\ \cline{3-4}

\multicolumn{1}{c|}{} & 
\multicolumn{1}{c|}{} & 
\multicolumn{1}{c|}{GPT family} &
\begin{tabular}[c]{@{}c@{}}
\cite{wang2024selective}, \cite{lu2022quark}
\end{tabular} \\ \cline{2-4}

\multicolumn{1}{c|}{} & 
\multicolumn{1}{c|}{\multirow{2}{*}{\makecell[c]{Copyright Protection}}} & 
\multicolumn{1}{c|}{LLaMA2} &
\begin{tabular}[c]{@{}c@{}}
\cite{eldan2023s}
\end{tabular} \\ \cline{3-4}

\multicolumn{1}{c|}{} & 
\multicolumn{1}{c|}{} & 
\multicolumn{1}{c|}{Diffusion Model} &
\begin{tabular}[c]{@{}c@{}}
\cite{wu2024erasediff}, \cite{gandikota2023erasing}, \cite{zhang2023forget}
\end{tabular} \\ \cline{2-4}

\multicolumn{1}{c|}{} & 
\multicolumn{1}{c|}{Hallucination Reduction} & 
\multicolumn{1}{c|}{MLLM} &
\begin{tabular}[c]{@{}c@{}}
\cite{liang2024unlearning}
\end{tabular} \\ \cline{1-4}

\multicolumn{1}{c|}{\multirow{2}{*}{\makecell[c]{\textcolor{blue}{\textbf{Knowledge Distillation}}}}} & 
\multicolumn{1}{c|}{\multirow{2}{*}{\makecell[c]{Privacy Compliance}}} &
\multicolumn{1}{c|}{GPT family/LLaMA2} &
\begin{tabular}[c]{@{}c@{}}
\cite{dong2024unmemorization}, \cite{huang2024offset}
\end{tabular} \\ \cline{3-4}

\multicolumn{1}{c|}{} & 
\multicolumn{1}{c|}{} &
\multicolumn{1}{c|}{DistillBERT, encoder-decoder transformers} &
\begin{tabular}[c]{@{}c@{}}
\cite{wang2023kga}
\end{tabular} \\ \cline{1-4}

\multicolumn{1}{c|}{\multirow{3}{*}{\makecell[c]{\textcolor{blue}{\textbf{Data Sharding}}}}} & 
\multicolumn{1}{c|}{Bias/Unfairness Alleviation} &
\multicolumn{1}{c|}{BERT/Distilgpt2} &
\begin{tabular}[c]{@{}c@{}}
\cite{kadhe2023fairsisa}
\end{tabular} \\ \cline{2-4}

\multicolumn{1}{c|}{} & 
\multicolumn{1}{c|}{\multirow{2}{*}{\makecell[c]{Privacy Compliance}}} &
\multicolumn{1}{c|}{LSTM} &
\begin{tabular}[c]{@{}c@{}}
\cite{liu2024forgetting}
\end{tabular} \\ \cline{3-4}

\multicolumn{1}{c|}{} & 
\multicolumn{1}{c|}{} &
\multicolumn{1}{c|}{Diffusion Model} &
\begin{tabular}[c]{@{}c@{}}
\cite{dai2023training}
\end{tabular} \\ \cline{1-4}

\multicolumn{1}{c|}{\multirow{6}{*}{\makecell[c]{\textcolor{blue}{\textbf{Extra Learnable Layers}}}}}  & 
\multicolumn{1}{c|}{\multirow{2}{*}{\makecell[c]{Safety Alignment}}} &
\multicolumn{1}{c|}{T5} &
\begin{tabular}[c]{@{}c@{}} 
\cite{chen2023unlearn}
\end{tabular} \\ \cline{3-4}

\multicolumn{1}{c|}{} & 
\multicolumn{1}{c|}{} &
\multicolumn{1}{c|}{Diffusion Models} &
\begin{tabular}[c]{@{}c@{}}
\cite{huang2023receler}
\end{tabular} \\ \cline{2-4}

\multicolumn{1}{c|}{} & 
\multicolumn{1}{c|}{\multirow{3}{*}{\makecell[c]{Privacy Compliance}}} &
\multicolumn{1}{c|}{T5} &
\begin{tabular}[c]{@{}c@{}}
\cite{chen2023unlearn}
\end{tabular} \\ \cline{3-4}

\multicolumn{1}{c|}{} & 
\multicolumn{1}{c|}{} &
\multicolumn{1}{c|}{BERT} &
\begin{tabular}[c]{@{}c@{}}
\cite{kumar2022privacy}
\end{tabular} \\ \cline{3-4}

\multicolumn{1}{c|}{} & 
\multicolumn{1}{c|}{} &
\multicolumn{1}{c|}{Diffusion Models} &
\begin{tabular}[c]{@{}c@{}}
\cite{huang2023receler}
\end{tabular} \\ \cline{2-4}

\multicolumn{1}{c|}{} & 
\multicolumn{1}{c|}{\multirow{1}{*}{\makecell[c]{Hallucination Reduction}}} &
\multicolumn{1}{c|}{GPT Family} &
\begin{tabular}[c]{@{}c@{}}
\cite{wang2024large}
\end{tabular} \\ \cline{1-4}

\multicolumn{1}{c|}{\multirow{4}{*}{\makecell[c]{\textcolor{blue}{\textbf{Task Vector}}}}}  & 
\multicolumn{1}{c|}{\multirow{3}{*}{\makecell[c]{Safety Alignment}}} &
\multicolumn{1}{c|}{LLaMA2/OPT} &
\begin{tabular}[c]{@{}c@{}} 
\cite{liu2024towards}, \cite{pochinkov2024dissecting}
\end{tabular} \\ \cline{3-4}

\multicolumn{1}{c|}{} &
\multicolumn{1}{c|}{} &
\multicolumn{1}{c|}{GPT family} &
\begin{tabular}[c]{@{}c@{}}
\cite{pochinkov2024dissecting}
\end{tabular} \\ \cline{3-4}

\multicolumn{1}{c|}{} &
\multicolumn{1}{c|}{} &
\multicolumn{1}{c|}{T5} &
\begin{tabular}[c]{@{}c@{}}
\cite{ilharco2022editing}
\end{tabular} \\ \cline{2-4}

\multicolumn{1}{c|}{} & 
\multicolumn{1}{c|}{Copyright Protection} & 
\multicolumn{1}{c|}{LLaMA3} &
\begin{tabular}[c]{@{}c@{}}
\cite{dou2024avoiding}
\end{tabular} \\ \cline{1-4}

\multicolumn{1}{c|}{\multirow{2}{*}{\makecell[c]{\textcolor{blue}{\textbf{PEMO}}}}}  & 
\multicolumn{1}{c|}{Safety Alignment} & 
\multicolumn{1}{c|}{LLaMA2/OPT/Alpaca} &
\begin{tabular}[c]{@{}c@{}}
\cite{zhou2023making}, \cite{hu2024separate}, \cite{zhang2024composing}
\end{tabular} \\ \cline{2-4}

\multicolumn{1}{c|}{} & 
\multicolumn{1}{c|}{Hallucination Reduction} &
\multicolumn{1}{c|}{LLaMA2/OPT} &
\begin{tabular}[c]{@{}c@{}}
\cite{hu2024separate}
\end{tabular} \\ \cline{1-4}

\multicolumn{1}{c|}{\multirow{2}{*}{\makecell[c]{\textcolor{red}{\textbf{In-Context Unlearning}}}}}  & 
\multicolumn{1}{c|}{Hallucination Reduction} & 
\multicolumn{1}{c|}{BERT/GPT family} &
\begin{tabular}[c]{@{}c@{}}
\cite{das2024larimar}
\end{tabular} \\ \cline{2-4}

\multicolumn{1}{c|}{} & 
\multicolumn{1}{c|}{Privacy Compliance} & 
\multicolumn{1}{c|}{T5} &
\begin{tabular}[c]{@{}c@{}}
\cite{pawelczyk2023context}
\end{tabular} \\ \cline{1-4}

\bottomrule[1.5pt]
\end{tabular}
}
\end{table}

\subsection{Parameter Optimization}
\subsubsection{ \textbf{Overview:}} The unlearning approach through parameter optimization aims to adjust specific model parameters to selectively unlearn certain behaviors without affecting other functions. This method aims to perform minimal alterations to parameters linked with target forget data $\tilde{\mathcal{D}}_f$ influences or biases to preserve essential model performance. However, such direct modifications to parameters can unintentionally impact unrelated parameters, potentially reducing model utility. Parameter optimization for unlearning can be implemented through various methods, including
\textbf{gradient-based} adjustments, \textbf{knowledge distillation}, \textbf{data sharding}, integrating \textbf{extra learnable layers}, \textbf{task vector}, and \textbf{parameter efficient module operation}.

\subsubsection{\textbf{Gradient Based}} Gradient based unlearning approaches aim to adjust the model parameters in a way that selectively forgets the knowledge associated with specific data points or patterns. This is achieved by optimizing the model's loss function in reverse (for gradient ascent) or forward (for gradient descent) directions to effectively remove or mitigate the learned associations without significantly impacting the model's overall performance.

\noindent\textbf{Gradient-based with reverse loss:}
The gradient ascent (GA) based approach is originate from a typical optimization-based technique raised by~\cite{thudi2022unrolling}. Given a target forget set $\tilde{\mathcal{D}_f}$ and an arbitrary loss function $\mathcal{L}(\theta)$, the GA algorithm iteratively update the model at each training step $t$:
\[\theta_{t+1} \leftarrow \theta_t + \lambda \nabla_{\theta_t}\mathcal{L}(\theta)\]
where $\lambda$ is the unlearning rate and $\theta_t$ denotes model parameters at step $t$. The objective of such approach reverts the change of the gradient descent without reverse loss term during the training with its opposite operation. 
\textbf{KUL}~\cite{jang2022knowledge} implements the gradient ascent (GA) method to intentionally maximize the loss function, thereby shifting the model's predictions for specific target samples in the opposite direction. Notably, during the unlearning process, KUL updates only a small subset of $\theta$ to unlearn the target unlearn sample $\tilde{\mathcal{D}_f}$. However, KUL may be impractical in scenarios where sensitive information is not explicitly defined. Building on this concept, \textbf{UnTrac}~\cite{isonuma2024unlearning} employs the GA approach but extends its application to estimate the influence of a training dataset on a test dataset through unlearning.
Nevertheless, though GA based approach can obtain an excellent unlearning performance, this usually comes with a large sacrifice of the model utility with non-target samples (i.e. catastrophic collapse), as it elaborated in~\cite{liu2024towards, yao2023large}. To address this utility degradation, \textbf{NPO}\cite{zhang2024negative} leverages the principles of preference optimization but uniquely applies it using only negative samples. By modifying the standard preference optimization to exclude positive examples, NPO aims to make the model forget target undesirable data $\tilde{\mathcal{D}_f}$ by decreasing their prediction probability. Additionally, the work theoretically shows that the progression toward catastrophic collapse can be exponentially slower when using NPO loss than pure GA. Specifically, NPO approach utilizes the following loss function:
\[
\mathcal{L}_{NPO, \beta}(\theta) = -\frac{2}{\beta} \mathbb{E}_{\mathcal{D}_f} \left[ \log \sigma \left(-\beta \log \frac{g^\ast(y|x)}{g(y|x)}\right) \right].
\]
In this formulation, \( \beta \) is a scaling parameter that adjusts the sharpness of the preference modifications. 
The term \( \log \frac{g^\ast(y|x)}{g(y|x)} \) measures how the predictions of the current unlearned model deviate from those of the pre-trained model. This loss function aims to minimize the likelihood of the model predicting outcomes linked to the \( \tilde{\mathcal{D}}_{f} \), effectively diminishing the model's recall of this data. 

Besides, numerous other works have built based on GA approach and extend with more efficient module. 
For example, \textbf{LLMU}~\cite{yao2023large} 
builds upon traditional GA methods and incorporates two arbitrary loss functions $\mathcal{L}_{\text{forget}}$ and $\mathcal{L}_{\text{mismatch}}$ to prevent model from generating unwanted outputs and maintains model functionality on normal prompts, which can be expressed as:
\[\theta_{t+1} = \theta_t - \epsilon_1 \cdot \nabla_{\theta_t} \mathcal{L}_{\text{forget}} - \epsilon_2 \cdot \nabla_{\theta_t} \mathcal{L}_{\text{mismatch}} - \epsilon_3 \cdot \nabla_{\theta_t} \mathcal{L}_{\text{maintain}}.
\] 
Similarly, \textbf{Eraser}~\cite{lu2024eraser} enhances the $\mathcal{L}_{\text{forget}}$ objective by adding random prefixes and suffixes to each query from $\mathcal{D}_r$ during training to simulate jailbreaking attacks, ensuring robust unlearning against varied prompts. It uses GPT-3.5 to extract entities from unwanted knowledge, which in this specific work refers to harmful responses, creating a dataset that helps the model retain general knowledge about entities in jailbreak prompts. Eraser also captures refusal responses from the original aligned model $\theta$ and trains the unlearned model $\theta_u$ to replicate these refusals, maintaining the model's safety alignment by rejecting harmful inquiries. Besides, GA approach can be employed on MLLMs to mitigate multimodal hallucination problems. For example, \textbf{EFUF} \cite{xing2024efuf} operates by initially assessing text-image congruence using the CLIP model, which scores the relevance of text descriptions to their corresponding images. Based on these scores, the framework identifies hallucinated content—descriptions that mention objects not present in the images.

\noindent\textbf{Gradient-based without reverse loss: }
Given that the gradient ascent (GA) based approach may degrade model performance, regular gradient descent offers a more balanced method for model training and unlearning. 
Within this framework, various innovative methods have been developed. 
Inspired by the Lottery Ticket Hypothesis~\cite{frankle2018lottery}, \textbf{PCGU}~\cite{yu2023unlearning} hypothesizes that a subset of the model's neurons encode biases/preferences in different contexts. PCGU introduces a weight importance algorithm using gradient calculations to rank weights by their contribution to biased outputs. Specifically, model parameter $\theta$ and gradient $\nabla_i$ are partitioned into $\{\theta^1, \theta^2, ... \theta^m\}$ and $\{\nabla_i^1, \nabla_i^2, ..., \nabla_i^m\}$. The impact of the weights on biased outputs is quantified by calculating the dot product $\theta^j \cdot \nabla_i^j$ for each $j$, identifying weights with the highest influence on biased behavior for targeted modification in the unlearning phase.

Building on the foundation of targeting specific model components for unlearning, \cite{gu2024second} proposes two novel unlearning methods: \textbf{Fisher Removal} and \textbf{Fisher Forgetting} for LLM leveraging second-order information, specifically the Hessian, to enhance the unlearning process. In particular, the Fisher Removal method updates the model parameters aggressively to ensure the removal of undesirable data, which can sometimes compromise the utility of the model. The objective of this method can be formulated as:
\[
\theta_{u} = \theta - \gamma \cdot F^{-1} \cdot \nabla L(\theta), 
\]
where $F^{-1}$ represents the inverse Fisher information. Then, Fisher Forgetting is a less aggressive variant of Fisher Removal, better preserving the model's accuracy even through multiple cycles of unlearning. This method modifies the parameters by adding a controlled amount of noise $\eta$:
\[
\theta_{u} = \theta - \gamma \cdot F^{-1} \cdot \nabla L(\theta) + \eta.
\]
Further advancing the complexity of unlearning strategies, \textbf{min-min optimization}~\cite{li2023make} adapts the bi-level optimization approach previously applied to images for creating unlearnable text, which the objective can be formulated as:
\[
\text{argmin}_{\theta} \mathbb{E}_{(x+\eta,y)\sim D} \left[ \text{argmin}_{\eta} \mathcal{L}(g(x+\eta), y) \right].
\]
Specifically, the method leverages error-minimization modifications to alter targeted forget data $\tilde{\mathcal{D}_f}$ through a controllable noise $\eta$ in such a way that it remains unlearnable even for models not seen during the optimization process. Building on gradient descent applications in machine learning, this approach adapts to generative models like vision models. \citet{li2024machine} introduces a framework for i2i (image-to-image) generative models using a two-part encoder-decoder architecture. The encoder converts input from $\tilde{\mathcal{D}_f}$ into a representation vector, while the decoder reconstructs the image. The objective is to maximize the output distribution difference between the retained data ($\mathcal{D}_r$) and the target forget set ($\tilde{\mathcal{D}}_f$), minimizing modifications to the retained data. Expanding on manipulating model outputs, \citet{kumari2023ablating} minimizes KL divergence between conditional distributions of target and anchor concepts, modifying the model to generate images of the anchor concept when prompted with the target concept. 

Besides manipulating the distance between $\tilde{\mathcal{D}}_f$ and $\mathcal{D}_r$, these datasets can be directly used for fine-tuning the model for unlearning. For instance, \textbf{UBT}~\cite{liang2024unlearning} mitigates backdoor attacks by dividing $\mathcal{D}_r$ into suspicious and clean subsets using a pre-trained model based on multimodal text similarity. Target unlearning samples ($\tilde{\mathcal{D}}_f$) are fine-tuned to amplify backdoor features, increasing cosine similarity measures. A token-level local unlearning strategy then selectively forgets tokens associated with backdoor triggers. Target unlearned samples $\tilde{\mathcal{D}}_f$ (i.e. negative samples) are also effective in vision generative models. \textbf{Forget-Me-Not}~\cite{zhang2023forget} uses curated images related to the concept targeted for removal during fine-tuning, adjusting the model's cross-attention layers within the UNet architecture to minimize attention to these concepts. \textbf{ESD} \cite{gandikota2023erasing} refines this by modifying model parameters to remove undesired concepts $\tilde{\mathcal{D}}_f$, such as nudity or copyrighted styles, without dataset censorship or post-generation filtering. ESD fine-tunes the model using a short textual description of the target undesired concepts ($\tilde{\mathcal{D}}_f$), aiming to reduce the probability of generating images described by $x$:
\[
\epsilon_{\theta_u}(y_t, x, t) \leftarrow \epsilon_{\theta_o}(y_t, t) - \eta[\epsilon_{\theta_o}(y_t, x, t) - \epsilon_{\theta^*}(y_t, t)]
\]
Negative sample fine-tuning may involve complexities in identifying undesirable behaviors accurately, leading to potential biases. Hence, some approaches directly use $\mathcal{D}_r$ for fine-tuning. For example, \textbf{EraseDiff} formulates unlearning as a constrained optimization problem. The objective is to fine-tune the model using the remaining data $\mathcal{D}_r$ to preserve utility while erasing the influence of forget data. This is achieved by deviating the generative process from the ground-truth denoising procedure:
\[
\min_{\theta_o} \mathcal{L} (\theta_o, \mathcal{D}_r) \quad \text{s.t.} \quad g(\theta_o, \tilde{\mathcal{D}}_f) - \min_{\phi|\theta_o} g(\phi, \tilde{\mathcal{D}}_f) \leq 0,
\]
where $g(\cdot)$ measures the influence of $\tilde{\mathcal{D}}_f$. The constraint ensures that unlearning targets $\tilde{\mathcal{D}}_f$ without degrading performance on $\mathcal{D}_r$.

Building on those fine-tuning approaches, which directly modify model behaviors using specific dataset, \textbf{Selective Amnesia (SA)}~\cite{heng2024selective} introduces a controllable forgetting mechanism applied to conditional variational likelihood models, including variational autoencoders (VAEs) and large-scale text-to-image diffusion models. SA utilizes continual learning techniques like Elastic Weight Consolidation (EWC) \cite{kirkpatrick2017overcoming} and Generative Replay (GR) \cite{shin2017continual} to preserve knowledge while selectively forgetting specific concepts. Additionally, SA introduces a surrogate objective to guarantee the reduction of the log-likelihood of the data to be forgotten, providing a more controlled and effective unlearning process. However, 
SA is more suitable for removing specific samples and fails to be generalizable to unlearning knowledge in a broader spectrum, such as nudity. 
Hence, \textbf{SalUn}~\cite{fan2023salun} method extends these principles to further refine the unlearning process by specifically targeting the modification of the model’s response to various class representations. In particular, the objective of concept unlearning for SalUn can be mathematically formulated as:
\[
\text{minimize} \quad L_{\text{SalUn}}(\theta_u) := \mathbb{E}_{(x,c) \sim \tilde{D_f}, t, \epsilon \sim \mathcal{N}(0,1), c' \neq c} \left[ \left\| \epsilon_{\theta_u}(x_t | c') - \epsilon_{\theta_u}(x_t | c) \right\|_2^2 \right] + \beta \mathcal{L}_{\text{MSE}}(\theta_u; D_r),
\]
where $\epsilon_{\theta_u}(x_t | c)$ and $\epsilon_{\theta_u}(x_t | c')$ represent the noise generator parameterized by $\theta_u$, conditioned on the text prompt $c$ and $c'$.

\subsubsection{\textbf{Knowledge Distillation}} Knowledge distillation approaches typically involve a teacher-student model configuration and treat the unlearned model $g^\ast$ as the student model, aiming to mimic the desirable behavior from a teacher model. 
\textbf{KGA}~\cite{wang2023kga} introduces a novel framework to align knowledge gaps, defined as differences in prediction distributions between models trained on different data subsets. The KGA framework adjusts the model to minimize the discrepancy in knowledge between a base model trained on all data and new models trained on an external set $\mathcal{D}_n$ similar to $\mathcal{D}$ but excluding $\mathcal{D}_f$. This is formulated as:
\[
f^* = \text{argmin}_f \left| \text{dis}_{(\mathcal{D}_n)}(g, g_n) - \text{dis}_{(\tilde{\mathcal{D}}_f)}(g, g_f) \right|.
\]
Here, \( g \) is the original model trained on \( \mathcal{D} \), while \( g_n \) and \( g_f \) are models trained on \(\mathcal{D}_n \) and \( \tilde{\mathcal{D}}_f \), respectively. The function \( \text{dis} \) measures the distributional differences using KL divergence. KGA treats the original model $g$ as a teacher model and minimizes the distance of output distributions when feeding samples in $\mathcal{D}_r$ to $g^\ast$ and $g$.
Moreover, \cite{dong2024unmemorization} introduces a novel LLM unlearning approach named \textbf{deliberate imagination (DI)} to further maintain model's generation and reasoning capabilities. DI employs a self-distillation technique in which a teacher model guides an LLM to generate creative and non-harmful responses, rather than merely forgetting unwanted memorized information. The process begins by strategically increasing the probability tokens within the teacher model that serve as alternatives to the memorized ones, thereby encouraging the production of novel and diverse outputs. Subsequently, DI fine-tunes the student model using the predicted output probabilities from the teacher, enabling the student models to generate imaginative responses that are less dependent on memorized data.


Similar to many white-box approaches, the KGA approach depends on accessing the model's internal weights, which makes it inapplicable to black-box models. To address this issue, \textbf{$\delta$ learning}~\cite{huang2024offset} introduces a novel framework for unlearning in black-box LLMs without accessing to the model’s internal weights. Specifically, $\delta$ learning utilizes a pair of smaller, trainable models to compute a logit offset that is then applied to the black-box LLM to adjust its responses. These smaller trainable models, referred to as offset models, are trained to predict how the logit outputs of the black-box LLM should be modified to exclude the influence of $\tilde{\mathcal{D}}_f$. The offset is calculated as the difference in logits between these two models and added to the black-box LLM's logits, guiding the final output away from sensitive information.

\subsubsection{\textbf{Data Sharding}} Data sharding approaches usually divide the training data into multiple shards, where each corresponds to a subset of the overall data $D$. Inspired by SISA~\cite{bourtoule2021machine}, separate models are trained for each data shards that can effectively remove target data in different shard based on request. As it mentioned in Section~\ref{sec:background}, SISA provides an exact unlearning guarantee due to the data to be forgotten have no impacts on the retrained version of the model, making it suitable to wide range of model architectures, including GenAI. However, direct implementation of SISA on GenAI is infeasible due to its the high computational cost in model saving, retraining, and inference. Hence, various work have adapted the similar approach of SISA and made more suitable to GenAI depends on its corresponding unlearning scenario. \textbf{FairSISA}~\cite{kadhe2023fairsisa} proposes a post-processing unlearning strategy for bias mitigation in ensemble models produced by SISA. In particular, this post-processing function \( f \) aims to enforce fairness in the outputs of \( g^* \), particularly ensuring that the model's predictions do not exhibit bias related to a sensitive attribute \( A \). This can be formulated as follows: \( f \) post-processes the outputs of \( g^* \) to satisfy fairness constraints, such as equalized odds, across groups defined by \( A \) without significantly affecting the overall model accuracy:
\[\min_{f} \mathbb{E}[\ell(f(g^*(x)), Y)]\]
subject to:
\[\Pr(f(g^*(x)) = 1 | A = 0, Y = y) = \Pr(f(g^*(x)) = 1 | A = 1, Y = y), \quad \forall y \in \{0, 1\}.\]
Here, \( \ell \) is a loss function, and \( \mathbb{E} \) denotes the expected value, reflecting the objective to minimize any deviation from the true labels \( Y \) while adhering to the fairness constraints.

Next, to reduce retraining costs while preserving utility during inference, \citet{liu2024forgetting} introduces the \textbf{LOO} ensemble method to unlearn target token sequences from LMs. This method uses a teacher-student framework where multiple teacher models are trained on partitioned data segments. When data removal is requested, the student model (i.e. base LM) is supervised by the remaining teachers, excluding the one trained on the segment to be unlearned. The recalibration is formalized as:
\[
g_{LOO-E}^{-k}(w|w_1^{t-1}) = \frac{1}{M-1} \sum_{m=1}^M \mathbbm{1}\{\mathcal{D}_k \not\subset B_m\} \cdot p_{\theta_m}(w|w_1^{t-1})
\]
where \( g_{LOO-E}^{-k} \) is the aggregated prediction excluding the teacher trained with \( \mathcal{D}_k \subset \mathcal{D}_f \). \( \theta_m \) denotes the parameters of the m-th teacher model, and \( w_1^{t-1} \) reflects the input sequence up to step \( t-1 \). The student model fine-tunes its parameters based on KL divergence, ensuring sensitive data is unlearned while maintaining performance on \( \mathcal{D}_r \). Despite improved utility performance compared to SISA, LOO lacks a theoretical guarantee for teacher models and may impose significant computational overheads on GenAI like LLMs.

The data sharding approach can be applied to vision generative models and LLMs for measuring training data influence for unlearning. \citet{dai2023training} presents \textbf{temporary unlearning} using ensemble ablation. The core idea is to use an ensemble of diffusion models, each trained on a subset of data, to assess the influence of individual data points by excluding models exposed to the data point and observing changes in outputs. The work provides a theoretical guarantee that the resulting model has not seen the removed training sample. The ensemble is represented as \( f_e \):
\[
f_e(x, t) = \mathbb{E}_{S \sim \mathscr{D}}[\mathbb{E}_{r \sim R}[f(x, t, A(S, r))]]
\]
where \( x \) is the input, \( t \) is the diffusion time, \( \mathscr{D} \) represents the uniform distribution over \( 2^{\mathcal{D}} \), and \( A(\cdot) \) denotes the training procedure with training samples and exogenous noise \( r \). To assess the influence of a specific point \( \tilde{x} \), the ensemble without \( \tilde{x} \), \( f_{e}^{-\tilde{x}} \), is evaluated as:
\[
f_{e}^{-\tilde{x}}(x, t) = \frac{1}{\text{Pr}(x \in S' \sim \mathcal{X})} \mathbb{E}_{S \sim \mathscr{D}}[\mathbb{E}_{r \sim R}[f(x, t, A(S, r)) \mathbbm{1}_{\tilde{x} \notin S}]].
\]
This method aligns with machine unlearning by providing a more efficient way to disregard specific data influences without extensive retraining. Using an ensemble approach ensures effective data attribution and compliance with privacy and fairness.

\subsubsection{\textbf{Extra Learnable Layers}}
Another parameter optimization approach is to introduce additional parameters or trainable layers in the model and train them to actively forget different sets of data from $\tilde{\mathcal{D}}_f$. This approach negates the necessity of modifying the model’s inherent parameters, thereby preventing interference to its original knowledge. \textbf{EUL}~\cite{chen2023unlearn} integrates an additional unlearning layer, \( U_l^{(i)} \), into the transformer following the feed-forward networks. Throughout the training, this unlearning layer is exclusively engaged to learn to forget the specified data, while the rest of the model parameters, denoted as \( \theta_o \), remain frozen. Upon receiving a deletion request, the model first trains a distinct unlearning layer \( U_l^{(i)} \) with parameters \( \theta_i \) tailored to that request. Subsequently, these layers are merged using a fusion mechanism to form a unified unlearning layer with parameters \( \theta_m \), achieved by minimizing a regression objective:
\begin{equation*}
\min_{\theta_m} \sum_i \left\| \theta_m^T X_f^{(i)} - \theta_i^T X_f^{(i)} \right\|^2
\end{equation*}
where \( \theta_m \) is defined as:
\begin{equation*}
\theta_m = \left( \sum_i X_f^{(i)T} X_f^{(i)} \right)^{-1} \sum_i \left( X_f^{(i)T} X_f^{(i)} \theta_i \right).
\end{equation*}
This results in a unified unlearning transformer capable of handling multiple deletion requests sequentially, ensuring that all specified data is effectively forgotten while preserving the integrity of model behaviors on other tasks.

Similarly, \textbf{Receler}~\cite{huang2023receler} introduces lightweight eraser parameters $\theta_{E}$ during unlearning, which is designed to remove the target concept from the outputs of each cross-attention layer within the diffusion U-Net. To erase the concept, the eraser is trained to predict the negatively guided noises that move the model’s prediction away from the erased concept. In particular, the objective is defined as:
\begin{align*}
\mathcal{L}_{\text{Erase}} &= \mathbb{E}_{y_t, t}\left[ \| \epsilon_{\theta_u}(y_t, e_x, t) - \epsilon_{E} \|^{2} \right]\\
\text{where } \epsilon_{E} &= \epsilon_{\theta}(y_t, t) - \eta \left[ \epsilon_{\theta}(y_t, e_x, t) - \epsilon_{\theta}(y_t, t) \right],
\end{align*}
where $\theta_u$ is the designated unlearned model with $\theta_o$ plugged with eraser $\theta_e$. $y_t$ is the denoised image at timestep $t$ sampled from $\theta_u$ conditioned on target concept $x$, $e_x$ is the text embedding of concept $x$, and $\epsilon_E$ is the negatively guided noise predicted by the original model $\theta_o$. By minimizing the L2 distance between $\epsilon_{\theta_u}(y_t, e_x, t)$ and $\epsilon_{E}$, the eraser learns to reduce the probability of the generated image $y$ belongs to the target concept $x$, thus effectively erasing the concept.

\citet{kumar2022privacy} proposes two extensions to the SISA framework \cite{bourtoule2021machine}— \textbf{SISA-FC} and \textbf{SISA-A} — to facilitate guaranteed unlearning that is efficient in terms of memory, time, and space for LMs. In particular, \textbf{SISA-FC} pre-trains a base model on a generic text corpus and then integrates fully connected layers atop it. During optimization, only the parameters of the linear layers are fine-tuned. This approach reduces the overall training time since backpropagation of gradients occurs solely in the final layers, and only the weights of the additional parameters are stored. Nevertheless, the addition of linear layers to SISA might compromise the model's utility when compared to fine-tuning the entire model \cite{devlin2018bert}. To address this, \textbf{SISA-A} incorporates adapters~\cite{houlsby2019parameter} into the encoder blocks of the transformer. This method results in only a marginal increase in the model's memory footprint—about 1 to 5 \% —thus providing a memory benefit of 95 to 99 \%.

\subsubsection{\textbf{Task Vector Methods}}
Inspired by recent work of weight interpolations ~\cite{frankle2020linear, wortsman2022robust, matena2022merging, ilharco2022patching, ainsworth2022git}, \citet{ilharco2022editing} first proposes the concept of task vector, which can be obtained by taking the difference between the original model weights of a pre-trained model and its weights fine-tuned on a specific task. In particular, if we let $\theta^{t}_{ft}$ be the corresponding weights after fine-tuning on task $t$, the task vector is then denoted as $\tau_t = \theta^{t}_{ft} - \theta_o$. Then, taking the element-wise negation of the task vector $\tau_t$ can enable $\theta_o$ to forget target knowledge on task $t$ without jeopardizing irrelevant knowledge, resulting in an unlearned model that has weight of $\theta_u = \theta_o - \lambda \tau$ with $\lambda$ as a scaling term. One exemplar work is \textbf{SKU}~\cite{liu2024towards}, which designs a novel unlearning framework to eliminate harmful knowledge while preserving utility on normal prompts. In particular, SKU is consisted of two stages where the first stage aims to identify and acquire harmful knowledge within the model, whereas the second stage targets to remove the knowledge using element-wise negation operation. Different from pure gradient ascent approaches where the model locality is largely compromised, SKU collectively aggregates the target unlearned knowledge $\tilde{\mathcal{D}}_f$ using gradient decent approach in the first stage and remove it from the pre-trained model. However, \textbf{SSU}~\cite{dou2024avoiding} identifies the potential instability of the pure task vector approach in the case of multiple rounds of unlearning (i.e., sequential unlearning) and introduces a more stable unlearning framework integrated with weight saliency.


Besides subtracting undesirable parameters, ~\citet{pochinkov2024dissecting} proposes a \textbf{selective pruning} method to trim those neurons' relative importance to different datasets, representing target model capability for unlearning. In particular, it performs either iterative pruning on nodes in the feed-forward layers or attention head layers. This selective approach utilizes importance functions that assess the contribution of individual neurons to specific tasks by measuring activation frequencies and magnitudes, enabling precise targeting of neurons that are crucial for the undesired capabilities. Furthermore, different from previous weight pruning method, where it requires Hessian computation, the selective neuron pruning is more computational efficient for large language models because it directly removes neurons that contribute the most to the unwanted behavior.

\subsubsection{\textbf{Parameter Efficient Module Operation Methods}}
Inspired by works on merging model parameters under full fine-tuning~\cite{wortsman2022model, matena2111merging, jin2022dataless}, \citet{zhang2024composing} explores composing parameter-efficient modules (PEM) like LoRA~\cite{hu2021lora} and $\text{(IA)}^3$~\cite{liu2022few} for flexible module manipulation. Unlike task vector methods, which modify the global weight vectors, PEM operation methods apply localized adjustments within specific modules. Similar to task vector methods, the negation operator ($\ominus$) in PEM methods unlearns stored knowledge within adapter modules. For example, in a LoRA module denoted as $\theta_{\text{lora}} = \{\mathbf{A}, \mathbf{B}\}$, $\mathbf{A}$ is initialized following a random Gaussian distribution, and $\mathbf{B}$ is initialized to all zeros to recover the pre-trained model at the beginning. We could negate $\mathbf{B}$ or $\mathbf{A}$ while keeping the other unchanged to facilitate unlearning or forgetting certain skills (e.g., toxic data). This process can be written as:
\[
\ominus \theta_{\text{lora}}^{\text{negation}} = \ominus \theta_{\text{lora}} = \{\mathbf{A}, -\mathbf{B}\}.
\]
This negation operator changes the intermediate layer's activation values in the opposite direction of gradient descent, aiding in unlearning targeted knowledge.

To further enhance model truthfulness and detoxification, \textbf{Ext-Sub}~\cite{hu2024separate} introduces "expert" and "anti-expert" PEMs. The "expert" PEM, trained on retaining data $\mathcal{D}_r$, represents desired behaviors, while the "anti-expert" PEM, trained on unlearned data $\tilde{\mathcal{D}}_f$, represents harmful behaviors. Ext-Sub identifies commonalities between these PEMs to determine shared capabilities and then subtracts the deficiency capability responsible for untruthful or toxic content. The Ext-Sub operation is defined as:
\[
\theta_u = \theta_{expert} \ominus \lambda \cdot \text{Ext} (\theta_{anti-expert}),
\]
where $\theta_{expert}$ and $\theta_{anti-expert}$ are the parameters of the expert and anti-expert PEMs, respectively, and Ext($\cdot$) is the extraction function isolating the deficiency capability. This process removes harmful effects while preserving and enhancing the beneficial capabilities of the LLM.

Additionally, PEM can be utilized during model training to prevent the acquisition of harmful information. For instance, \citet{zhou2023making} introduces \textbf{security vectors}, enabling LLMs to be exposed to harmful behaviors without modifying the model's original, safety-aligned parameters. Specifically, these vectors allow the LLM to process and respond to harmful inputs during training, ensuring that the core parameters remain unaltered. The security vector $\theta_s$ is optimized to minimize the loss associated with harmful data:
\[\arg \min_{\theta_o} \mathbb{E}_{(X,Y) \sim D_{u}} \left[ \min_{\theta_s} L(f(X; \theta_o; \theta_s), Y) \right],
\]
where $L$ is the causal loss computed based on the prediction and the ground truth. Subsequently, the trained security vector $\theta^{\ast}_{s}$ is employed during fine-tuning to ensure that the model learns from benign data without adopting harmful behaviors. $\theta^{\ast}_{s}$ is activated during the forward pass of training to maintain the model's response consistency with benign data and to inhibit the learning of harmful information. Specifically:
\[\theta_u = \arg\min_{\theta_o} \mathbb{E}_{(X,Y) \sim D_{r}} \left[ L\left(f(X; \theta_o; \theta_s^*), Y\right) \right],\]
ensuring that the original model parameters do not update in a harmful direction.

\subsubsection{ \textbf{Summary:}} 

The parameter optimization strategies focus on adjusting specific model parameters to selectively unlearn certain behaviors without affecting other functions. These approaches involve precise alterations to parameters associated with unwanted data influences or biases, ensuring the preservation of essential model performance. Gradient-based approaches with reversed loss are effective for unlearning accuracy and generalizability but can negatively impact model locality by inadvertently affecting unrelated parameters. In contrast, gradient-based methods without reversed loss can maximally preserve locality but may not excel in unlearning accuracy and generalizability. Extra learnable layers provide highly targeted unlearning but may demand significant computational resources. Data sharding methods excel in maintaining locality by partitioning the training data and ensuring specific data points can be unlearned without extensive retraining, although they might struggle with generalizability in very large models. Knowledge distillation is effective in maintaining locality by transferring knowledge to a new model trained to exclude specific data, thus retaining essential performance while unlearning undesired knowledge. However, it can be resource-intensive and may not achieve satisfactory accuracy and generalizability. Task vector and parameter-efficient module operations may perform well in terms of unlearning accuracy and generalizability. Nonetheless, recent work \cite{dou2024avoiding} has highlighted the risk of these approaches leading to instability due to significant model degradation, resulting in poor locality performance.

\subsection{In-Context Unlearning}

\subsubsection{\textbf{Overview }}
Unlike parameter optimization approaches, which actively modifies parameters either locally or globally via different techniques, in-Context unlearning techniques retain the parameters in their original state and manipulate the model's context or environment to facilitate unlearning. In particular, these strategies result in an unlearned model $g^{\ast}_{\theta_u}$ where $\theta_u = \theta_o$, but with changes in how the model interacts with its input or inferences. 

\subsubsection{\textbf{In-Context Unlearning}}
In-context unlearning utilizes the approach of in-context learning to selectively erase targeted knowledge during inference, treating the model as a black box. This kind of method is resource-efficient but has inherent limitations due to the nature of in-context learning. Specifically, it modifies only the model's immediate outputs without fundamentally eradicating the unwanted knowledge embedded within the model's internal parameters. 
\textbf{ICUL}~\cite{pawelczyk2023context} first introduces the idea of in context unlearning, which alters input prompts during the inference phase to achieve targeted unlearning in the situation where model's API is the only access to the model. The technical process involves several key steps: 
\begin{enumerate}
    \item \textbf{Label Flipping}, where the label of the data point that needs to be forgotten is flipped to contradict the model’s learned associations, resulting in the template "$[Forget \> Input]_0$ $[Flipped \> Label]_0$"
    \item \textbf{Context Construction}, where additional correctly labeled examples are sampled and appended to the flipped example, creating a mixed input sequence, resulting in the template "$ [Forget \> Input]_0 \ [Flipped \> Label]_0 \ \backslash n; [Input]_1 \; [Label]_1 \ \backslash n; \ldots \; [Input]_s \; [Label]_s$
    \item \textbf{Inference Adjustment}, where this modified prompt is used to 'confuse' the model about the original training point, mitigating its influence, resulting in the template "$[Forget \> Input]_0$ $[Flipped \> Label]_0$ $\backslash n; [Input]_1 \>; [Label]_1 \> \backslash n; \ldots \>; [Input]_s \>; [Label]_s\>  [Query\> Input]_{s+1}$".    
\end{enumerate}
However, directly modifying input prompts does not always give a desirable output. Hence, \textbf{Larimar}~\cite{das2024larimar} integrates an external memory module that directly manipulates the LLM's outputs. This approach allows for precise control over knowledge updates and selective forgetting through operations such as writing, reading, and generating, which are efficiently managed by a hierarchical memory system. 

\subsubsection{\textbf{Summary}} Parameter frozen methods retain the model's parameters in their original state while manipulating the model's context or environment to facilitate unlearning. A notable advantage of this method is its resource efficiency, as it does not require retraining or modification of the model's internal parameters, making it suitable for scenarios where direct access to the model's internals is limited (e.g. black-box models). However, a significant drawback is that parameter frozen methods only modify the model's immediate outputs without fundamentally eradicating the unwanted knowledge embedded within the model's parameters. This can lead to incomplete unlearning, as the underlying knowledge remains intact.

\section{Datasets and Benchmarks}
\label{sec:datasets}

\subsection{Datasets}
In this section, we summarize the datasets commonly used in the field of Generative AI, as outlined in Table~\ref{tab:dataset}, to benefit future MU research. Instead of merely categorizing the datasets by task (i.e., generation and classification), they are organized according to their intended unlearning objectives. We specifically focus on those datasets primarily utilized as target datasets during the unlearning process, excluding object removal datasets such as CIFAR10 and MNIST, as well as generic evaluation benchmark datasets like Hellaswag~\cite{zellers2019hellaswag} and Piqa~\cite{bisk2020piqa}.

\begin{table}[!t]
\centering
\caption{
Representative statistics of datasets for GenAI MU approaches with different generative types, including datasets for both \textcolor{blue}{\textbf{generation (G)}} and \textcolor{red}{\textbf{classification (C)}} tasks.
}\vspace{-0.1in}
\renewcommand{\arraystretch}{1.2}
\setlength{\tabcolsep}{3.pt}
\setlength{\aboverulesep}{0pt}
\setlength{\belowrulesep}{0pt}
\scalebox{0.8}{
\begin{tabular}{l|l|l|l|l|l}
\midrule[1pt]
\textbf{Dataset} & \textbf{Application} & \textbf{GenAI Type} & \textbf{Task} & \textbf{Instance} & \textbf{Used in} \\ 
\hline
LAION & \multicolumn{1}{c|}{\multirow{4}{*}{\makecell[c]{Safety Alignment}}} & \multicolumn{1}{c|}{\multirow{1}{*}{\makecell[c]{Gen Image Models}}} & \textcolor{blue}{\textbf{G}} & 5.85 B & \cite{gandikota2023erasing, zhang2023forget}\\
\cline{3-6}
Civil Comments & \multicolumn{1}{c|}{} & \multicolumn{1}{c|}{\multirow{3}{*}{\makecell[c]{(L)LMs}}} & \textcolor{red}{\textbf{C}} & 1,999,514 & \cite{ilharco2022editing}\\
PKU-SafeRLHF & \multicolumn{1}{c|}{} & \multicolumn{1}{c|}{} & \textcolor{blue}{\textbf{G}} & 834,000 & \cite{yao2023large, liu2024towards}\\
Anthropic red team & \multicolumn{1}{c|}{} & \multicolumn{1}{c|}{} & \textcolor{blue}{\textbf{G}} & 38,961 & \cite{zhou2023making}\\
\hline
Harry Potter & \multicolumn{1}{c|}{\multirow{3}{*}{\makecell[c]{Copyrights Protection}}} & \multicolumn{1}{c|}{\multirow{3}{*}{\makecell[c]{(L)LMs}}} & \textcolor{blue}{\textbf{G}} & 7 books & \cite{yao2023large, thaker2024guardrail, eldan2023s}\\
Bookcorpus & \multicolumn{1}{c|}{} & \multicolumn{1}{c|}{} & \textcolor{blue}{\textbf{G}} & 11,038 books & \cite{yao2023large, lu2022quark, isonuma2024unlearning, yao2024machine}\\
TOFU & \multicolumn{1}{c|}{} & \multicolumn{1}{c|}{} & \textcolor{blue}{\textbf{G}} & 200 profiles & \cite{thaker2024guardrail, zhang2024negative}\\
\hline
HaluEVAL & \multicolumn{1}{c|}{\multirow{5}{*}{\makecell[c]{Hallucination Reduction}}} & \multicolumn{1}{c|}{\multirow{4}{*}{\makecell[c]{(L)LMs}}} & \textcolor{blue}{\textbf{G}} & 35,000 & \cite{yao2023large}\\
TruthfulQA & \multicolumn{1}{c|}{} & \multicolumn{1}{c|}{} & \textcolor{blue}{\textbf{G}} & 817 & \cite{yao2023large, isonuma2024unlearning, liu2024towards}\\
CounterFact & \multicolumn{1}{c|}{} & \multicolumn{1}{c|}{} & \textcolor{blue}{\textbf{G}} & 21,919 & \cite{das2024larimar, wang2024large}\\
ZsRE & \multicolumn{1}{c|}{} & \multicolumn{1}{c|}{} & \textcolor{blue}{\textbf{G}} & Unknown & \cite{das2024larimar, wang2024large}\\
\cline{3-6}
MSCOCO & \multicolumn{1}{c|}{} & \multicolumn{1}{c|}{\multirow{1}{*}{\makecell[c]{MLLMs}}} & \textcolor{blue}{\textbf{G}} & 328,000 & \cite{gandikota2024unified, gandikota2023erasing}\\
\hline
Pile & \multicolumn{1}{c|}{\multirow{9}{*}{\makecell[c]{Privacy Compliance}}} & \multicolumn{1}{c|}{\multirow{7}{*}{\makecell[c]{(L)LMs}}} & \textcolor{blue}{\textbf{G}} & Unknown & \cite{zhao2023learning, isonuma2024unlearning, wang2024selective, dong2024unmemorization}\\
Yelp/Amazon Reviews & \multicolumn{1}{c|}{} & \multicolumn{1}{c|}{} & \textcolor{red}{\textbf{C}} & 142.8 M & \cite{pawelczyk2023context}\\
SST-2 & \multicolumn{1}{c|}{} & \multicolumn{1}{c|}{} & \textcolor{blue}{\textbf{G}} & 11,855 & \cite{zhang2024composing, kumar2022privacy, pawelczyk2023context, li2023make}\\
PersonaChat & \multicolumn{1}{c|}{} & \multicolumn{1}{c|}{} & \textcolor{blue}{\textbf{G}} & 8784 & \cite{wang2023kga}\\
LEDGAR & \multicolumn{1}{c|}{} & \multicolumn{1}{c|}{} & \textcolor{blue}{\textbf{G}} & 1,081,177 & \cite{wang2023kga}\\
SAMSum & \multicolumn{1}{c|}{} & \multicolumn{1}{c|}{} & \textcolor{blue}{\textbf{G}} & 16,369 & \cite{chen2023unlearn, zhao2023learning}\\
IMDB & \multicolumn{1}{c|}{} & \multicolumn{1}{c|}{} & \textcolor{red}{\textbf{C}} & 50,000 & \cite{chen2023unlearn}\\
\cline{3-6}
CelebA-HQ & \multicolumn{1}{c|}{} & \multicolumn{1}{c|}{\multirow{2}{*}{\makecell[c]{Gen Image Models}}} & \textcolor{blue}{\textbf{G}} & 30,000 & \cite{panda2023fast, wu2024erasediff, dai2023training, moon2024feature}\\
I2P & \multicolumn{1}{c|}{} & \multicolumn{1}{c|}{} & \textcolor{blue}{\textbf{G}} & 4703 & \cite{gandikota2023erasing, wu2024erasediff, huang2023receler}\\
\hline
StereoSet & \multicolumn{1}{c|}{\multirow{3}{*}{\makecell[c]{Bias/Unfairness Alleviation}}} & \multicolumn{1}{c|}{\multirow{3}{*}{\makecell[c]{(L)LMs}}} & \textcolor{blue}{\textbf{G}} & 17,000 & \cite{yu2023unlearning}\\
HateXplain & \multicolumn{1}{c|}{} & \multicolumn{1}{c|}{} & \textcolor{blue}{\textbf{G}} & Unknown & \cite{kadhe2023fairsisa}\\
CrowS Pairs & \multicolumn{1}{c|}{} & \multicolumn{1}{c|}{} & \textcolor{blue}{\textbf{G}} & 1508 & \cite{yu2023unlearning}\\

\midrule[1pt]
\end{tabular}}
\label{tab:dataset}
\end{table}

\subsubsection{Safety Alignment: }

The \textbf{Civil Comments} dataset~\cite{borkan2019nuanced} is comprised of public comments from various news websites, each labeled with a level of toxicity to represent the degree of harmfulness in the content. Studies such as~\cite{ilharco2022editing} and \cite{zhang2024composing} have utilized subsets of these highly toxic samples to extract harmful knowledge from pre-trained models. Complementing this, the \textbf{Anthropic red team} dataset~\cite{bai2022training, ganguli2022red} includes human preference data and annotated dialogues for evaluating red team attacks on language models. This dataset aids in reducing harm in generative models, as demonstrated by ForgetFilter~\cite{zhao2023learning}, which classifies conversations into safe and unsafe categories to facilitate unlearning harmful responses, and by \cite{zhou2023making}, which generates security vectors to address harmful knowledge.

Next, the \textbf{PKU-SafeRLHF} dataset~\cite{ji2024beavertails} consists of 30,000 expert comparison entries with safety meta-labels, initially proposed by BeaverTails to train moderation models preventing LLMs from generating harmful outputs. This dataset has been instrumental in various studies; for instance, LLMU \cite{yao2023large} leverages harmful samples for gradient ascent, while SKU~\cite{liu2024towards} extracts harmful knowledge from multiple perspectives in pre-trained models. Lastly, moving to the multimodal domain, the \textbf{LAION} dataset~\cite{schuhmann2022laion} contains 5.85 billion CLIP-filtered image-text pairs, supports large-scale multi-modal model training. Despite its democratizing potential, it poses challenges, such as the generation of NSFW content by diffusion models. To mitigate these risks, recent studies like~\cite{zhang2023forget} and \cite{gandikota2023erasing} have identified and targeted subsets of NSFW data within LAION for content removal.

\subsubsection{Copyrights Protection: }

The \textbf{Harry Potter}~\cite{rowling1997harry} has been a focus for studies aiming to eliminate a model's ability to generate Harry Potter-related content while preserving performance. \cite{eldan2023s} achieved this by fine-tuning the model with a dataset where idiosyncratic expressions were replaced with common counterparts. Similarly, \cite{yao2023large} transformed the book into a question-and-answer format to evaluate the model's content generation post-unlearning. However, due to copyright protections, this dataset is not publicly available. In contrast, the \textbf{BookCorpus} dataset~\cite{Zhu_2015_ICCV}, a large collection of freely available novels, comprises 11,038 books across various sub-genres. It was used in training the initial GPT model by OpenAI~\cite{radford2018improving} and has been utilized in GenAI unlearning. \cite{yao2023large} used BookCorpus as a retaining dataset to maintain performance on unrelated samples, while UnTrac~\cite{isonuma2024unlearning} employed it to assess the impact of dataset removal.

\subsubsection{Hallucination Eliminations: }

The \textbf{HaluEVAL} dataset~\cite{li2023halueval} comprises 5,000 general user queries with ChatGPT responses and 30,000 task-specific examples across question answering, knowledge-grounded dialogue, and text summarization. This dataset is ideal for hallucination elimination due to its intentionally misleading samples, serving as targets for unlearning~\cite{yao2023large}. In contrast, the \textbf{CounterFact} dataset~\cite{meng2022locating} includes 21,919 factual relations with prompts and paired true and false responses, making it suitable for examining how models process and recall factual knowledge. Larimar~\cite{das2024larimar} utilizes this dataset to test its ability to replace incorrect facts with correct ones during unlearning, assessing the model’s precision in updating its episodic memory. Meanwhile, the \textbf{TruthfulQA} dataset~\cite{lin2021truthfulqa} assesses whether a language model produces truthful answers to 817 questions across 38 categories, including health, law, finance, and politics. LLMU \cite{yao2023large} initially used TruthfulQA to maintain model utility while unlearning hallucinations. TruthfulQA is also considered generic knowledge that the model should preserve \cite{liu2024towards, isonuma2024unlearning}. Finally, the \textbf{MS COCO} dataset~\cite{lin2014microsoft}is a comprehensive collection for object detection, segmentation, key-point detection, and captioning, comprising 328,000 images. EFUF~\cite{xing2024efuf} leverages this dataset to extract objects from generated captions and calculate their image relevance using CLIP~\cite{radford2021learning}, aiding in unlearning hallucinations in multimodal models. The detailed annotations make it suitable for object/concept unlearning in vision generative models~\cite{gandikota2023erasing}.

\subsubsection{Privacy Compliance: }
The \textbf{Pile}~\cite{pile} is a comprehensive, open-source language modeling dataset aggregating 825 GiB from 22 diverse, high-quality datasets. This compilation significantly enriches data diversity, enhancing general cross-domain knowledge and downstream generalization capabilities of large-scale models. ForgetFilter~\cite{zhao2023learning} utilizes subsets of The Pile, post-processed with toxicity scores, to unlearn harmful information. Similarly, studies such as \cite{dong2024unmemorization}, DeMem~\cite{kassem2023preserving}, \cite{jang2022knowledge}, and \cite{wang2024selective} use the Training Data Extraction Challenge—a set of 15,000 easily extractable samples from The Pile's training set—to simulate privacy unlearning requests. \cite{pochinkov2024dissecting} also leverages The Pile as both forget and retain datasets under different settings to unlearn multiple skills from pre-trained LLMs.

Transitioning to sentiment analysis, the \textbf{Yelp/Amazon Reviews} datasets offer extensive collections of user-generated content from Amazon.com and Yelp.com, respectively, including detailed reviews, ratings, and metadata about products and businesses. These datasets are important to privacy compliance; for instance, ICUL~\cite{pawelczyk2023context} evaluated privacy-preserving algorithms using 25,000 data points from each dataset, focusing on minimizing information leakage while maintaining robust sentiment classification performance. Similarly, the \textbf{SST-2} dataset~\cite{socher-etal-2013-recursive} comprises 11,855 sentences from movie reviews, parsed using the Stanford parser and fully labeled with parse trees. This dataset facilitates an in-depth analysis of compositional sentiment effects in language. ICUL~\cite{pawelczyk2023context} and \cite{zhang2024composing} selected subsets from SST-2 as target data for unlearning, evaluating the effectiveness of their proposed unlearning algorithms. In the context of sentiment classification, the \textbf{IMDB} dataset~\cite{maas2011learning} includes user reviews of movies, directors, actors, and more, categorized into two sentiment types. The trained model must disregard comments related to randomly selected movies or individuals from the training set during testing. 

Moving to dialogue data, \textbf{PersonaChat}~\cite{zhang2018personalizing} features crowd-sourced dialogues where each speaker bases their conversation on a provided profile, making it suitable for unlearning sensitive information. KGA~\cite{wang2023kga} utilizes the official train, validation, and test splits to conduct experiments on this dataset. For legal text classification, \textbf{LEDGAR}~\cite{tuggener2020ledgar} consists of legal provisions in contracts sourced from SEC filings, with over 12,000 labels annotated across nearly 100,000 provisions in more than 60,000 contracts. KGA~\cite{wang2023kga} uses a subset of this dataset for privacy information unlearning experiments. The \textbf{SAMsum} dataset~\cite{gliwa-etal-2019-samsum} contains approximately 16,000 English conversations designed to mimic everyday messaging, complete with summaries for each conversation. This dataset is useful for evaluating unlearning in conversational models.

For vision tasks, the \textbf{CelebA-HQ} dataset~\cite{karras2017progressive} features more than 200K celebrity images with 40 attribute annotations, suitable for tasks like face attribute recognition and landmark localization. It is employed as a target dataset for feature-level unlearning in vision generative models like GANs and VAEs~\cite{moon2024feature, tiwary2023adapt}. The \textbf{I2P} dataset~\cite{schramowski2023safe} includes real user prompts from text-to-image tasks designed to evaluate and mitigate inappropriate imagery in vision generative models. Receler~\cite{huang2023receler}, \cite{gandikota2023erasing}, and EraseDiff~\cite{wu2024erasediff} utilize these prompts to generate and unlearn inappropriate images in diffusion models.

\subsubsection{Bias/Unfairness Alleviation: }
The \textbf{StereoSet} dataset~\cite{nadeem2020stereoset} is designed to measure stereotype bias in language models, consisting of 17,000 sentences that evaluate model preferences across gender, race, religion, and profession. To perform well on StereoSet, an unlearned model must demonstrate fairness and unbiased behavior while possessing a strong comprehension of natural language. PCGU~\cite{yu2023unlearning} utilizes the intrasentence subset of StereoSet to assess its effectiveness in unlearning biases, particularly by adjusting model parameters to reduce biased predictions and enhance the fairness of its outputs. Similarly, \textbf{CrowS-Pairs} dataset~\cite{nangia2020crows} is a crowdsourced collection specifically designed to measure the extent of stereotypical biases present in large pre-trained masked language models. This dataset comprises 1,508 examples that address stereotypes related to nine types of bias, including race, religion, and age. Each example pairs two sentences: one that stereotypes a historically disadvantaged group and another that less stereotypically represents an advantaged group, with minimal differences in wording between the two. This setup makes it suitable for unlearning existing biases in pre-trained models. In PCGU~\cite{yu2023unlearning}, the authors evaluate masked language models for bias by comparing the probabilities associated with each sentence. 

Additionally, the \textbf{HateXplain} dataset~\cite{mathew2021hatexplain} serves as a benchmark for explainable hate speech detection, focusing on Twitter and Gab posts annotated by Amazon Mechanical Turk workers. It uniquely includes word and phrase level annotations that capture human rationales for each labeling decision, categorizing posts as hate speech, offensive, or normal. In FairSISA~\cite{kadhe2023fairsisa}, the authors utilize these detailed annotations to ensure fair and sensitive handling of race and religion, aggregating subgroups into broader categories to address the sparse data distribution and overlap among sensitive attributes.

\subsection{Benchmarks}
In the subsequent, we provide detailed explanations of various benchmarks for different generative models to assist in evaluating their unlearning performance from a variety of perspectives.

\subsubsection{Vision Generative Model}

\textbf{UnlearnCanvas~\cite{zhang2024unlearncanvas}} is a benchmark designed to evaluate how diffusion models can forget certain styles or objects they’ve learned. For example, a model can be asked to erase Van Gogh’s influence from their digital brushstrokes. UnlearnCanvas evaluates these generative image models by challenging them to drop specific styles or objects from their repertoire without affecting their overall artistic talent. In particular, it does this by feeding them with a variety of images, from sketches to watercolors, and observing how well they can keep generating while forgetting what we ask them to.

\subsubsection{Large Language Model}

The \textbf{TOFU} benchmark~\cite{maini2024tofu} is designed to evaluate the unlearning capabilities of large language models, specifically focusing on the models' ability to forget specific data about fictitious authors. It provides a controlled environment with 200 synthetic author profiles to test how effectively models can eliminate selected information while retaining other data. TOFU employs a suite of metrics to measure both the residual knowledge of forgotten content and the overall utility of the model post-unlearning, challenging models to demonstrate selective forgetting without compromising their performance on unrelated tasks. Next, \textbf{WMDP} benchmark~\cite{li2024wmdp} aims to identify and mitigate the risk of LLMs aiding in the development of biological, cyber, and chemical weapons. It introduces a public dataset of over 4,000 multiple-choice questions as a proxy to measure hazardous knowledge in these areas. The WMDP benchmark serves two main purposes: to evaluate LLMs for dangerous knowledge that could enable malicious use, and to test unlearning methods capable of removing such knowledge.

\subsubsection{Multimodal (Large) Language Model }

The \textbf{Object HalBench}~\cite{rohrbach2018object} is commonly used to evaluate object hallucination in detailed image descriptions by comparing the objects generated in model outputs against comprehensively annotated object labels for COCO images~\cite{lin2014microsoft}. Building on this, the \textbf{MMHal-Bench}~\cite{sun2023aligning} focuses on assessing hallucinations and the informativeness of responses, utilizing GPT-4 to compare the model's output with human responses and a range of object labels. However, due to incomplete text annotations in MMHal-Bench, it is primarily used to measure the model's level of informativeness. The \textbf{MHumanEval} benchmark~\cite{yu2023rlhf} encompasses both long-form image descriptions and short-form questions, containing 146 samples gathered from both Object HalBench and MMHal-Bench. Human annotators label the hallucinated segments and the types of hallucinations (e.g., objects, positions, numbers) based on the model responses. Furthermore, \textbf{LLaVA Bench}~\cite{liu2024visual} is widely utilized for evaluating the helpfulness of multimodal conversations, focusing on detailed descriptions and complex reasoning capabilities, with helpfulness scores reported based on reference responses from GPT-4. Lastly, the \textbf{POPE} benchmark~\cite{li2023evaluating} evaluates MLLMs by checking if they hallucinate any objects not present in the images. It uses simple yes-or-no questions about object presence, varying the objects between random, commonly seen, and those likely to cause errors, thus offering a scalable and clear way to measure models' reliability in accurately recognizing image content.

\section{Applications}
\label{sec:applications}

GenAI MU can benefit multiple downstream applications with the ability of precisely yet efficiently removing undesirable knowledge from pre-trained generative models. In the following, we introduce several key applications of generative model MU techniques in realistic situations. The detailed downstream applications with different approaches can be found in Table \ref{tab:approach}.

\subsection{Safety Alignment:}
\noindent\textbf{Existing Works.} Generative models have demonstrated remarkable proficiency in generating text, images, and multimodal outputs. However, the diversity and vastness of their pre-trained data can lead to safety concerns. 
Thus, MU emerges as an effective tool for generative models to mitigate the influence of undesirable data while preserving the model's utility for relevant tasks. The extensive use of Generative AI brings various safety concerns, including the generation of inappropriate content. Inappropriate content generation refers to the production of harmful, disturbing, or illegal outputs in response to problematic prompts or inputs. For example, parameter optimization approaches \cite{liu2024towards, ilharco2022editing, zhou2023making, zhang2024composing, pochinkov2024dissecting} are effective for safety alignment applications, as they reduce a model's harmful generation by subtracting the fine-tuned weights on harmful datasets from the pre-trained models while preserving the model's utility. Additionally, eradicating NSFW (Not Safe For Work) content, such as nudity, has become a critical focus in the field of unlearning generative image models \cite{zhang2023forget, fan2023salun, wu2024erasediff, gandikota2023erasing}. Models trained on large datasets like LAION \cite{schuhmann2022laion} may contain inappropriate content that can mislead the model, making this an important area of study.

\subsection{Privacy Compliance:}
\noindent\textbf{Existing Works.} Generative models raise privacy concerns due to their powerful memorization capabilities, which may result in data leakage during generation. This issue is exacerbated when models are trained on data collected from individual users, such as medical data \cite{d2022preserving}. Consequently, some users may request the model owner to delete their data in compliance with the Right To Be Forgotten (RTBF), as stipulated by recent privacy legislation~\cite{rosen2011right, hoofnagle2019european, pardau2018california}. Upon request, data samples containing sensitive information can be safely removed from the generative models using unlearning approaches. In practice, many works \cite{li2023make, chen2023unlearn, wang2024selective, huang2024offset, jang2022knowledge} have focused on introducing effective unlearning approaches to enhance privacy compliance. For example, KGA \cite{wang2023kga} proposes a comprehensive unlearning pipeline based on knowledge gap alignment. This method minimizes the knowledge gap between the output distributions of teacher-student models, enabling them to produce similar predictions for both seen and unseen data. DeMem \cite{kassem2023preserving} fine-tunes the model using a negative similarity score as a reward signal, encouraging the model to adopt a paraphrased policy that effectively forgets sensitive data from the pre-trained model. Additionally, Forget-Me-Not \cite{zhang2023forget} introduces a plug-and-play module to fine-tune the UNet in the Stable Diffusion model. This module uses a curated dataset to redirect the model's attention away from the target concept, such as a photograph of a person, thereby achieving unlearning in a more targeted and efficient manner.

\subsection{Copyrights Protection: }
\noindent\textbf{Existing Works.} As data becomes an increasingly critical component of generative models, the actual source data owners—who are the parties or individuals that own the originality of the works (e.g., books, artworks, and images, etc.)—raise significant concerns about the abusive usage of their data. Their data can be coincidentally or intentionally collected by the model trainer as pre-trained data samples to construct high-quality generative models. Additionally, possible data duplication may severely jeopardize the ownership of the original data samples. Hence, one of the unlearning objectives of generative models is to safely remove the copyrights of those unauthorized works to offer better ownership protection. The objective of protecting copyrights is very similar to that of privacy compliance, as both aim to prevent the unauthorized use of sensitive information. However, copyright protection is often more challenging due to the intricate nature of intellectual property laws and the diverse range of copyrighted materials involved. These two objectives overlap in the context of generative image models, where copyright can be regarded as a type of concept, such as the style of Van Gogh. Consequently, many approaches design techniques that are applicable to both objectives \cite{wang2024selective, kumari2023ablating, gandikota2023erasing, zhang2023forget}. However, the situation becomes increasingly complicated when applied to language models and multimodal (large) language models. For example, \cite{eldan2023s} proposes unlearning information related to \textit{Harry Potter} from a fine-tuned LLM via word replacement with generic translations, which may degrade model utility and raise the issue of hallucinations. A detailed analysis of unlearning copyrights can be found in Section \ref{sec: chal_copyright_unlearn}.

\subsection{Hallucinations Reduction: }
\noindent\textbf{Existing Works.} Given a fact-related question or the task of interpreting an informative image, generative models—especially Large Language Models (LLMs) and Multimodal (Large) Language Models (MLLMs)—often generate plausible yet false or incorrect responses, potentially misleading users \cite{yu2023rlhf, lin2021truthfulqa, liu2023mitigating}. Thus, unlearning could serve as a strategy to minimize such hallucinations in generative models by enabling them to forget incorrect answers and the associated misleading knowledge. This also includes some inconsistent, or outdated knowledge exists in the pre-training data corpus. As emphasized by LLMU \cite{yao2023large}, the goal of unlearning hallucinations is to prevent the model from providing incorrect answers rather than ensuring it always gives factually correct answers. The difference between unlearning and model editing for the seek of hallucination is highlighted in Section \ref{background: vs_editing}. Hence, negative sampling and contrastive-based approaches are typically implemented during the unlearning process. For instance, LLMU introduces a gradient-ascent-based approach to make the model learn in the opposite direction of the hallucinated answer. Similarly, UBT \cite{liang2024unlearning} proposes an approach to mitigate backdoor attacks in MLLMs by dividing the dataset into suspicious (i.e., hallucinated) and clean subsets, fine-tuning to amplify backdoor features, and then implementing a token-level local unlearning strategy to selectively forget identified backdoor triggers.

\subsection{Bias/Unfairness Alleviation: }
\noindent\textbf{Existing Works.} Previous studies~\cite{bender2021dangers, lee2021deduplicating} have shown that pre-training corpora contain many biased data samples. Neural models trained on such data may easily capture these biases and exhibit them during implementation. In particular, the negative stereotypes and biases embedded in the models may lead to potential unfairness and harm in a variety of applications, like medical care \cite{schmidgall2024addressing}, question-answer chatbots \cite{wan2023biasasker} and video/image generation \cite{cho2023dall}. Unlearning becomes an effective tool to mitigate this unfairness without massive retraining.  
Recently, several works have proposed to apply MU techniques in bias/unfairness alleviation of generative models. For example, \textbf{FairSISA}~\cite{kadhe2023fairsisa} proposes a post-processing bias mitigation strategy for ensemble models produced by SISA, aiming to enforce fairness in the outputs by ensuring that the model's predictions do not exhibit bias related to a sensitive attribute. In \textbf{PCGU}~\cite{yu2023unlearning}, the author introduce a weight importance algorithm that uses gradient calculations to rank and prioritize model weights based on their contribution to biased outputs, targeting the most influential weights for modification in the unlearning phase.

\section{Discussion}
\label{sec:discussion}

\subsection{Challenges}
\subsubsection{\textbf{Copyright Unlearning.}} 
\label{sec: chal_copyright_unlearn}

With the broad adoption of generative AI across various domains, the emphasis on copyright protection has significantly increased, necessitating that generative models eliminate detailed information about specific copyrighted works. Compared to generative image models, the task of copyright unlearning presents more challenges for language models due to the complexity of the issue and the ambiguity of the ultimate goal. Unlike unlearning targets like sensitive information or harmful knowledge—where the goal is to remove a particular sample and its influence without compromising the model's utility on unrelated samples—copyrighted products like books often cannot be easily separated from other pre-trained knowledge, particularly when it comes to specific sections. Additionally, it remains uncertain whether an adversarial attack could succeed by targeting a work similar to the copyrighted one. Recent efforts \cite{yao2023large, eldan2023s, dou2024avoiding} have attempted to unlearn copyright products by generating random responses or replacing sensitive words on large language models. However, these approaches can compromise model utility and exacerbate model hallucination, highlighting the need for improvements in both the objectives of unlearning copyright and the design of unlearning responses.

\subsubsection{\textbf{Theoretical Analysis.}}
While many current studies on generative model MU concentrate on developing effective methodologies to enhance unlearning performance for various targets, there remains a substantial gap between practical unlearning applications and theoretical analysis. Recent work by NPO~\cite{zhang2024negative} theoretically demonstrates that progression toward catastrophic collapse via minimizing the NPO loss occurs exponentially slower than with a Gradient Ascent (GA)-based approach in Large Language Models (LLMs). In general, incorporating theoretical analysis can enrich research from multiple perspectives and divide into two main streams. Firstly, it can offer theoretical insights to refine the objectives of generative model unlearning and its potential ties to differential privacy. Secondly, it can establish theoretical guarantees, like NPO~\cite{zhang2024negative}, to validate the efficacy of certain methods and their capability to balance intricate trade-offs. Both directions contribute to a more profound understanding of MU's underlying mechanisms, promoting more robust and systematic research within the field. Nonetheless, as the field of generative model MU is nascent, comprehensive theoretical analyses or guarantees tailored to generative models' MU challenges are yet to be developed. We hope that future studies will expand theoretical discussions and provide substantial contributions to the foundational understanding of generative model unlearning strategies.

\subsubsection{\textbf{Knowledge Entanglement}}
Knowledge entanglement presents yet another critical challenge to generative model unlearning, hindering the exclusion of specific information without influencing related knowledge or the model's overall functionality. The authors of TOFU~\cite{maini2024tofu} observe that models often forget unrelated information during unlearning. This issue arises from the blurred lines and overlap between concepts within the model's learning process. Such overlaps can lead to overgeneralization during unlearning, negatively affecting the model's performance on tasks that were not intended to be unlearned. For instance, ~\cite{jang2022knowledge} introduces knowledge unlearning to decrease privacy risks in language models after training, but its success can vary across data domains due to similarities among them. This issue also exists in generative image models, where unlearning specific classes or styles may inadvertently alter unrelated ones, as shown by \cite{gandikota2023erasing}. Therefore, knowledge entanglement poses a significant challenge to generative model unlearning due to its possible impacts on accuracy and scalability.

\subsubsection{\textbf{Knowledge Entanglement between Model Locality and Generalizability}}
In traditional Machine Unlearning (MU), numerous studies~\cite{liu2023breaking, chien2022efficient, guo2019certified, pan2023unlearning} have focused on mitigating the trade-off between unlearning effectiveness and model utility, which is also the goal for generative model MU. While some work like SKU~\cite{liu2024towards} have focused on alleviating the trade-off between unlearning efficacy and utility in LLM unlearning, these efforts primarily address the unlearning of harmful knowledge, and their generalizability to other unlearning targets remains unverified. For instance, FairSISA~\cite{kadhe2023fairsisa} highlights the challenges of balancing model accuracy with fairness enhancement, particularly when employing post-processing techniques that neither alter the training data nor the model directly. Furthermore, this trade-off is not exclusive to LLMs as other generative models face similar dilemmas. For example,~\cite{gandikota2023erasing} illustrate the need to balance complete erasure of specific concepts against minimal interference with other visual concepts, underlining an inherent trade-off in concept-specific unlearning. Similarly, the existing works on vision generative model~\cite{sun2023generative, huang2023receler, gandikota2023erasing} have also shown that the models often struggle to isolate the target concept without impacting related or co-occurring features. This interdependence can lead to model degradation, where the overall quality of generated outputs diminishes, or the model fails to preserve its generative capabilities for non-targeted concepts. Thus, tackling this trade-off and achieving a balance between the effectiveness of unlearning and the preservation of model utility remains a significant challenge for MU methods for generative models.

\subsection{Future Directions}
Despite the recent achievements in the development of Machine Unlearning (MU) approaches in the field of generative models, these methods are still in their nascent stages compared to traditional MU techniques. Therefore, in this section, we explore several promising directions that could be pursued or investigated to further advance the field.

\subsubsection{\textbf{Consistency of unlearning Targets.}} 
Due to constant knowledge updates, current GenAI MU approaches may not be consistent, meaning they might fail to erase newly introduced undesirable knowledge. For example, although ~\cite{zhou2023making} attempts to use security vectors to prevent models from acquiring specific types of knowledge, their work is not generalizable for removing newly introduced concepts during model updates.
In practice, it is very likely that models unlearned for certain knowledge may need to be fine-tuned on similar unlearned knowledge for different time steps. Therefore, a promising future direction involves maintaining the consistency of unlearning targets even after extensive knowledge updates.

\subsubsection{\textbf{Robust Unlearning.}} 
A critical direction for advancing GenAI MU lies in enhancing its robustness. LLMs are known to be vulnerable to jailbreak attacks~\cite{wei2024jailbroken, qi2023fine, huang2023catastrophic} and other attacks like backdoor attacks. Recently, this vulnerability has been shown in multimodal (large) language models as well~\cite{niu2024jailbreaking, gu2024agent, wu2023jailbreaking}. It is important to exploit unlearning as an effective solution to develop a defense pipeline to avoid GenAI from attacking. Various strategies for GenAI MU
have focused on improving the method's robustness against adversarial attacks. For instance, Eraser~\cite{lu2024eraser} develops an robust unlearning framework that aims to remove harmful knowledge without being compromised by various jailbreak attack methods. UnlearnDiffAtk~\cite{zhang2023generate} introduces an effective adversarial prompt generation method for diffusion models that simplifies the generation of adversarial prompts by reducing its reliance on auxiliary models. However, in an era where attack strategies develop rapidly, it is imperative that MU approaches for generative models not only retain their effectiveness but also demonstrate robustness against such attacks. For instance, a recent adaptive attack strategy~\cite{andriushchenko2024jailbreaking} has been able to jailbreak various state-of-the-art LLM after safety alignment, including GPT-4, Claude, and Gemini, with success rates close to 100\%. Therefore, more effort must be devoted to enhancing the robustness of unlearning algorithms to prevent various adversarial attacks.

\subsubsection{\textbf{Reliability of LLMs as Evaluators in GenAI Unlearning}}
Since human evaluators are typically expensive and time-consuming, an increasing number of studies and benchmarks have utilized state-of-the-art Large Language Models (LLMs) like GPT-4 as tools to assess unlearning performance and utility. For instance, SKU~\cite{liu2024towards} and \cite{zhou2023making} employ GPT-4~\cite{achiam2023gpt} and GPT-3.5-turbo respectively to evaluate the harmfulness of model responses post-unlearning. Similarly, Eraser~\cite{lu2024eraser} uses GPT-3.5 to derive a harmfulness score ranging between 1 and 5 to aid in determining the success rate of jailbreak attacks. Additionally, EFUF~\cite{xing2024efuf} utilizes GPT-4 to assess the informativeness of responses generated by the unlearned model to ensure that utility is not compromised after unlearning. Despite the prevalence and effectiveness of this approach, its validity is questionable. Firstly, LLMs are known to produce biased decisions and outputs~\cite{lucy2021gender, perez2022discovering, tamkin2023evaluating, cui2023holistic}, which may adversely affect the evaluation results. Secondly, many recent studies~\cite{lester2021power, webson2021prompt, sahoo2024systematic} have highlighted the importance of prompt templates during prompt tuning, which is predominantly used to calibrate GPT as an evaluator. This suggests that evaluation results from such an evaluator may largely depend on its system prompts and relevant templates, introducing uncertainties in the evaluation process. Hence, despite the challenges, it will be intriguing to see how future research addresses the drawbacks of using LLMs as evaluators in GenAI unlearning, aiming to provide more reliable and convincing evaluation outcomes.




\section{Conclusions}
\label{sec:conclusions}


In this survey, we present a comprehensive analysis of machine unlearning (MU) techniques in the domain of generative artificial intelligence (GenAI), encompassing generative image models, (large) language models (LLMs), and multimodal (large) language models (MLLMs). By categorizing existing unlearning methods into parameter optimization, and in-context unlearning strategies, we highlight the unique techniques, advantages, and limitations of each approach. Subsequently, we summarize the datasets and benchmarks widely utilized to measure GenAI techniques, emphasizing that different techniques may need to be assessed from different perspectives. Furthermore, to promote future researchers to investigate the field in diverse directions, we also identify the real-world applications of GenAI techniques. Meanwhile, we recognize several potential challenges and promising future directions to further enhance the efficiency and effectiveness of MU methods. Ultimately, this survey aims to deepen the understanding of MU in GenAI and inspire further advancements in the field to ensure safer and more trustworthy AI systems.

\bibliographystyle{ACM-Reference-Format}
\bibliography{acmart}

\end{document}